\documentclass{article}

\usepackage{microtype}

\usepackage{graphicx}
\usepackage{booktabs} 
\usepackage{lipsum}

\usepackage[breaklinks]{hyperref}

\usepackage[accepted]{icml2024}

\usepackage{amsmath}
\usepackage{amssymb}
\usepackage{mathtools}

\usepackage{amsthm}

\usepackage{multirow}       

\usepackage{url}
\usepackage[utf8]{inputenc} 
\usepackage[T1]{fontenc}    
\usepackage[breaklinks]{hyperref}      
\usepackage{url}            

\usepackage{booktabs}       
\usepackage{amsfonts}       
\usepackage{amsmath}
\usepackage{amssymb}
\usepackage{mathtools}
\usepackage{amsthm}
\usepackage{nicefrac}       
\usepackage{microtype}      
\usepackage{xcolor}         
\usepackage{wrapfig}
\usepackage{graphicx}
\usepackage{caption}
\usepackage{multirow,makecell}

\usepackage{xspace}
\usepackage{subcaption}

\usepackage{algorithm}
\usepackage{algorithmic}

\usepackage[capitalize,noabbrev]{cleveref}

\theoremstyle{plain}

\theoremstyle{definition}

\theoremstyle{remark}

\usepackage[textsize=tiny]{todonotes}

\icmltitlerunning{Extreme LLM Compression via Additive Quantization}

\begin{document}

\twocolumn[
\icmltitle{Extreme Compression of Large Language Models via Additive Quantization}



\icmlsetsymbol{equal}{*}

\begin{icmlauthorlist}

\icmlauthor{\hspace{5px}Vage Egiazarian}{equal,hse,yandex}
\icmlauthor{\hspace{-3px}Andrei Panferov}{equal,hse,yandex}
\icmlauthor{\hspace{-3px}Denis Kuznedelev}{yandex,skoltech}
\icmlauthor{\hspace{-3px}Elias Frantar}{istaustria}
\icmlauthor{\hspace{-3px}Artem Babenko}{yandex}
\icmlauthor{\hspace{-3px}Dan Alistarh}{istaustria,neuralmagic}

\end{icmlauthorlist}

\icmlaffiliation{hse}{HSE University}
\icmlaffiliation{yandex}{Yandex Research}
\icmlaffiliation{skoltech}{Skoltech}
\icmlaffiliation{istaustria}{IST Austria}
\icmlaffiliation{neuralmagic}{NeuralMagic}

\icmlcorrespondingauthor{}{dan.alistarh@ist.ac.at}
\newcommand{\methodname}{AQLM}

\icmlkeywords{Machine Learning, Quantization}

\vskip 0.3in
    ]



\printAffiliationsAndNotice{\icmlEqualContribution}

\begin{abstract}
The emergence of accurate open large language models (LLMs) has led to a race towards performant 
quantization techniques which can enable their  execution on end-user devices. 
In this paper, we revisit the problem of ``extreme'' LLM compression---defined as targeting extremely low bit counts, such as 2 to 3 bits per parameter---from the point of view of classic methods in Multi-Codebook Quantization (MCQ). 
Our algorithm, called AQLM, generalizes the classic \emph{Additive Quantization (AQ)} approach for information retrieval to advance the state-of-the-art in LLM compression, via two innovations: 1) learned additive quantization of weight matrices in input-adaptive fashion, and 2) joint optimization of codebook parameters across each transformer blocks. 
Broadly, AQLM is the first scheme that is Pareto optimal in terms of accuracy-vs-model-size when compressing to less than 3 bits per parameter, and significantly improves upon all known schemes in the extreme compression (2bit) regime. 
In addition, AQLM is practical: we provide fast GPU and CPU implementations of AQLM for token generation, which enable us to match or outperform optimized FP16 implementations for speed, while executing in a much smaller memory footprint. 
\end{abstract}

\section{Introduction}
\label{sect:intro}

\begin{figure}[ht]
    \centering
    \includegraphics[width=0.8\linewidth]{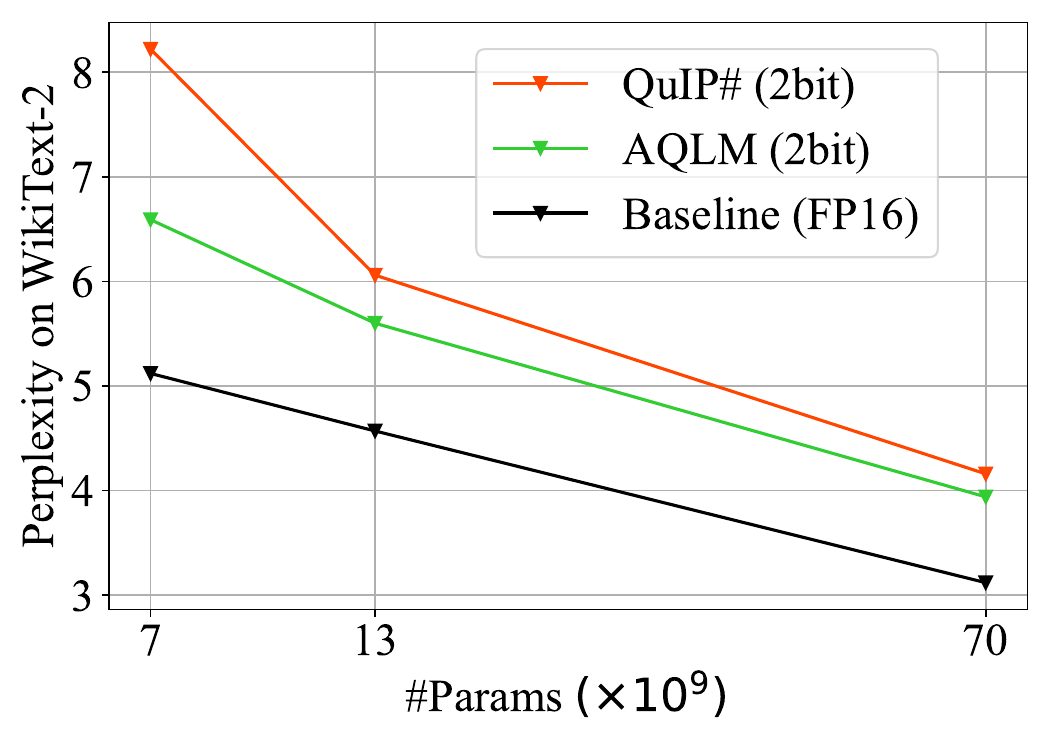}
    \caption{
    Comparison of AQLM (2-bit) relative to the state-of-the-art QuIP\# (2-bit) and the original 16-bit weights on \textsc{Llama 2} 7, 13, and 70B models.}
    \vspace{-15px}
    \label{fig:aqlm_vs_quip_sharp_2_bit}
\end{figure}

The rapid advancement of generative large language models (LLMs) has led to massive industrial and popular interest, driven in part by the availability of accurate \emph{open} LLMs, 
such as \textsc{Llama} 1 and 2~\cite{touvron2023llama}, Falcon~\cite{falcon2023}, BLOOM~\cite{scao2022bloom}, OPT~\cite{zhang2022opt}, or NeoX/Pythia~\cite{biderman2023pythia}.
A key advantage of open models is that they can be inferenced or fine-tuned  locally by end-users, assuming that their computational and memory costs can be reduced to be manageable on commodity hardware. 
This has led to several methods for inference and fine-tuning 
on compressed LLMs~\cite{dettmers2022llm, frantar2022gptq, dettmers2022case, lin2023awq, dettmers2023qlora}. Currently, the primary approach for accurate post-training compression of LLMs is \emph{quantization}, which reduces the bit-width at which model weights (and possibly activations) are stored, leading
to improvements in model footprint and
memory transfer.




By and large, LLM weights are compressed via ``direct'' quantization, in the sense that a suitable quantization grid and normalization are first chosen for each matrix sub-component, and then weights are each mapped onto the grid either by direct rounding, e.g.~\cite{dettmers2022case}, or via more complex allocations, e.g.~\cite{frantar2022gptq}. 
Quantization induces a natural compression-vs-accuracy trade-off, usually measured in terms of model size vs model perplexity (PPL). 
Existing approaches can achieve arguably low accuracy loss at 3-4 bits per element~\cite{dettmers2023spqr, chee2023quip, kim2023squeezellm}, and can even stably compress models to 2 or even less bits per element, in particular, for extremely large models \cite{frantar2023qmoe}. Yet, in most cases, low bit counts come at the cost of significant drops in accuracy, higher implementation complexity and runtime overheads. 
Specifically, from the practical perspective, ``extreme'' quantization in the 2-bit range using current techniques is inferior to simply using a smaller base model and quantizing it to higher bitwidths, such as 3-4 bits per parameter, as the latter yields higher accuracy given the same model size in bytes~\cite{dettmers2022case,chee2023quip}. 

{\bf Contribution.} In this work, we improve the state-of-the-art in LLM compression by showing for the first time that  \emph{Multi-Codebook Quantization (MCQ)} techniques can be extended to LLM weight compression. Broadly, MCQ is a family of information retrieval methods~\cite{rvq,pq,opq,compositeq,aq,lsq,lsq++}, consisting of specialized quantization algorithms to compress databases of vectors, allowing for efficient search.
Unlike direct quantization, MCQ compresses multiple values jointly, by leveraging the mutual information of quantized values.

More precisely, we extend Additive Quantization (AQ)~\cite{aq,lsq}, a popular MCQ algorithm, to the task of compressing LLM weights such that the output of each layer and Transformer block are approximately preserved. Our extension reformulates the classic AQ optimization problem to reduce the error in LLM layer outputs under the input token distribution and as well as to jointly optimize codes over layer blocks, rather than only preserving the weights themselves as in standard AQ. We refer to the resulting procedure as \emph{Additive Quantization of Language Models (AQLM)}. 
 Unlike some extreme LLM quantization approaches that require hybrid sparse-quantized formats which separate outlier quantization~\cite{kim2023squeezellm, dettmers2023spqr}, AQLM quantizes models in a simple homogeneous format, which is easy to support in practice. 
 Our main contributions are as follows: 
 \begin{enumerate}
    \vspace{-5px}\item We propose the AQLM algorithm, which extends AQ to post-training compression of LLM weights, via two innovations: 
    (1)  adapting the MAP-MRF\footnote{Maximum
a Posteriori inference in Markov Random Fields} optimization problem  behind AQ
    to be instance-aware, taking layer calibration input \& output activations into account;
    (2) complementing the layer-wise optimization with an efficient intra-block tuning technique, which optimizes quantization parameters jointly over several layers, using only the calibration data. 
    \item We evaluate the effectiveness of this algorithm on the task of compressing accurate open LLMs from the \textsc{Llama 2}~\cite{touvron2023llama} family with compression rates of 2-4 bits per parameter. We find that AQLM outperforms the previous state-of-the-art  across  the standard 2-4 bit compression range, with the most significant improvements for extreme 2-bit quantization (see Figure~\ref{fig:aqlm_vs_quip_sharp_2_bit}). 
    We provide detailed ablations for the impact of various algorithm parameters, such as code width and number of codebooks, and extend our analysis to the recent Mixtral model~\cite{jiang2024mixtral}. We also evaluate AQLM with improved fine-tuning algorithms from subsequent works, which leads to further increase in accuracy for 2- and 3-bit models.
    \item We show that AQLM is practical, by providing efficient GPU and CPU kernels implementations for specific encodings, as well as end-to-end generation\footnote{\url{https://github.com/Vahe1994/AQLM}}. Results show that our approach can match or even outperform the floating point baseline in terms of  speed, while reducing the memory footprint by up to 8x. 
    Specifically, AQLM can be executed with layer-wise speedups of $\sim 30\%$ for GPUs, and of up to 4x for CPU inference.

\end{enumerate}




\section{Background \& Related Work}

\subsection{LLM Quantization}\label{sect:related_llm}

Early efforts towards post-training quantization (PTQ) methods~\citep{nagel2020up, gholami2021survey} that scale to LLMs such as ZeroQuant~\citep{yao2022zeroquant}, LLM.int8()~\citep{dettmers2022llm}, and nuQmm~\citep{park2022nuqmm} employed direct round-to-nearest (RTN) projections, and adjusted quantization granularity to balance memory efficiency and accuracy. GPTQ~\citep{frantar2022gptq} proposed a more accurate \emph{data-aware approach} via an approximate large-scale solver for minimizing layer-wise $\ell_2$ errors.

\citet{dettmers2022case} examined the accuracy-compression trade-offs of these early methods, suggesting that 4-bit quantization may be optimal for RTN quantization, and observing that data-aware methods like GPTQ allow for higher compression, i.e. strictly below 4 bits/weight,  maintaining Pareto optimality. Our work brings this Pareto frontier below 3 bits/weight, for the first time. 
Parallel work quantizing both weights \emph{and activations} to 8-bits, by~\citet{dettmers2022llm},~\citet{xiao2022smoothquant}, and~\citet{yao2022zeroquant} noted that the ``outlier features'' in large LLMs cause substantial errors, prompting various mitigation strategies.

Recently, several improved techniques have focused on the difficulty of quantizing weight outliers, which have high impact on the output error. 
SpQR~\citep{dettmers2023spqr} addresses this by saving outliers as a highly-sparse higher-precision matrix. 
 AWQ~\citep{lin2023awq} reduces the error of quantizing channels with the highest activation magnitudes by employing per-channel scaling to reduce the error on important weights. SqueezeLLM~\citep{kim2023squeezellm} uses the diagonal Fisher as a proxy for the Hessian and implements non-uniform quantization through K-means clustering. 

 The published state-of-the-art method is  QuIP~\citep{chee2023quip}. Concurrent to our work, an improved variant called QuIP\#~\cite{quip-sharp} was introduced. Roughly, they work by first ``smoothening'' weights by multiplying with a rotation matrix, and then mapping them onto a lattice. 
At a high level, QuIP and QuIP\# aim to minimize the ``worst-case'' error for each layer, given initial weights and calibration data. For instance, in QuIP\#, the distribution of the rotated weights approximates a Gaussian, while the encoding lattice (E8P) is chosen to minimize ``rounding'' error. 
By contrast, our approach uses a different weight encoding (codebooks are  \emph{additive}), and \emph{learned} codebooks instead of a fixed codebook. Thus, our insight is that we should be able to obtain higher accuracy by \emph{direct optimization} of the codebooks over the calibration set, removing the rotation. Further, we show that codebooks for different layers can co-train via joint fine-tuning over the calibration data.

\subsection{Quantization for Nearest Neighbor Search}\label{sect:related_NNS}  

Our work builds on approximate nearest neighbor search (ANN) algorithms. Unlike PTQ,  ANN quantization aims to compress a database of vectors to allow a user to efficiently compute similarities and find nearest neighbors relative to a set of query points.
For high compression, modern ANN search algorithms employ \textit{vector quantization} (VQ)---which quantizes multiple vector dimensions jointly~\cite{vq1,vq2}.
It achieves this by learning ``codebooks'': i.e. a set of learnable candidate vectors that can be used to encode the data.
To encode a given database vector, VQ splits it into sub-groups of entries, then encodes every group by choosing a vector from the learned codebook. The algorithm efficiently computes distances or dot-products for similarity search by leveraging the linearity of dot products.

Quantization methods for ANN search generalize
vector quantization and are referred to as multi-codebook quantization (MCQ). MCQ methods typically do not involve information loss on the query side, which makes them the leading approach for memory-efficient ANN~\cite{competitveq, lsq++}. We briefly review MCQ  below.

\textbf{Product quantization (PQ)}~\cite{pq} is an early version of MCQ, which encodes each vector $x \in \mathbf{R}^D$ as a concatenation of $M$ codewords from $M$ $\frac{D}{M}$\nobreakdash-dimensional codebooks $C_1,\ldots,C_M$, each containing $K$ codewords. PQ decomposes a vector into $M$ separate subvectors and applies vector quantization (VQ) to each subvector, while using a separate codebook. Thus, each vector $x$ is encoded by a tuple of codeword indices $[i_1,\ldots,i_M]$ and approximated by $x \approx [c_{1i_1},\ldots, c_{Mi_M}]$.
Fast Euclidean distance computation becomes possible using lookup tables:

\vspace{-10px}\begin{gather}
\label{eq:adc}
\|q - x\|^2 \approx \|q - [c_{1i_1},\ldots, c_{Mi_M}]\|^2 = 
\sum\limits_{m=1}^{M}{\|q_m - c_{mi_m}\|}^2, \nonumber
\end{gather}\vspace{-5px}

where $q_m$ is the $m$th subvector of a query $q$. This sum can be calculated using $M$ additions and lookups if the distances from query subvectors to codewords are precomputed.
Since product-based approximations work better if the $\frac{D}{M}$-dimensional components independent distributions, subsequent work has  looked into finding better transformations \cite{opq, norouzi13}. As for the other similarity functions, \cite{guo2016quantization} proposes a quantization procedure for maximum inner product search (MIPS).
They minimize quantization error in the inner products between database and query vectors by solving a constrained optimization problem.
Similarly to the formula above, this procedure allows for efficient inner product search by precomputing dot products between the query $q$ an all codes in the learned codebooks, then adding these partial dot products to recover the full similarity score.


\textbf{Non-orthogonal quantizations.} Follow-up work \cite{rvq, aq, lsq, compositeq, competitveq, lsq++} generalized the idea of Product Quantization by approximating each vector by a \emph{sum} of $M$ codewords instead of concatenation. The resulting procedure is still efficient while the approximation accuracy is increased. 

\vspace{-2px}For this, Residual Vector Quantization~\cite{rvq}, quantizes original vectors, and then iteratively quantizes the approximation residuals from the previous iteration. Additive Quantization (AQ) \cite{aq} is more general, as it does not impose constraints on the codewords from the different codebooks. Usually, AQ provides the smallest compression errors, but is more complex to train for large $M$. 
We discuss this in detail in Section~\ref{sect:method}.

\vspace{-2px}Finally, several recent works~\cite{lsq,lsq++,compositeq} elaborate the idea of Additive Quantization, proposing the more effective procedure for codebooks learning. Composite Quantization (CQ) \cite{compositeq} learns codebooks with a fixed value of inner product between the codewords from different codebooks. Currently, the state-of-the-art compression accuracy is achieved by the LSQ method~\cite{lsq++}.

\paragraph{Vector quantization for model compression.} 
There has been significant work on exploiting vector quantization in the context of machine learning. 
For instance,~\citet{Zhou2017, li2017performance, Chen_Wang_Pan_2019} use multi-codebook quantization to compress word embeddings within deep learning models. 
Another line of work~\cite{maddness2021, McCarter2022LookupsAN,  FernndezMarqus2023AreWT} explores vector quantization for linear models, or linear layers within deep models. Similarly to PQ above, these techniques pre-compute inner products between inputs and all codes, then compute linear layer via look-up, which speeds up inference. However, these algorithms introduce significant prediction error that does not allow them to compress deep models. 
Thus, we believe we are the first to successfully adapt and scale MCQ to LLMs.


\section{AQLM: Additive Quantization for LLMs}\label{sect:method}



\subsection{Overview}

We start from the observation that additive quantization (AQ) solves a related problem to post-training quantization (PTQ)~\cite{nagel2020up, frantar2022obc}:  both settings assume the existence of a set of ``input'' vectors, i.e. input data for AQ, and the weight matrix rows for PTQ. The goal is to compress these inputs while preserving dot product similarity, against query vectors (for AQ), and against layer input embeddings (for PTQ).  
The difference between the two is that AQ assumes that the distribution of queries is unknown, whereas PTQ methods, e.g.~\cite{frantar2022obc}, show that it is sufficient to optimize for  sample input embeddings from a set of calibration data.

At a high level, we start by solving the following problem: for a linear layer
with $d_{in}$ input and $d_{out}$ output features
given its weights $\mathbf{W} \in \mathbb{R}^{d_{out} \times d_{in}}$ and a set of calibration inputs $\mathbf{X} \in \mathbb{R}^{d_{in} \times n}$, one seeks for a configuration of quantized weights $\widehat{\mathbf{W}}$ that optimizes squared error between the output of the original and compressed layer:

\vspace{-10px}\begin{equation}
    \underset{\widehat{\mathbf{W}}}
    {\arg\min}||\mathbf{W} \mathbf{X} - \widehat{\mathbf{W}} \mathbf{X} ||_2^2.
    \label{eq:obc}
\end{equation}\vspace{-10px}



In the following, we will assume that $\widehat{\mathbf{W}}$ is quantized using AQ, and adopt standard notation~\cite{lsq}. 
AQ splits weight rows into groups of $g$ consecutive elements, and represents each group of weights as a sum of $M$ vectors chosen from multiple learned codebooks $C_1, ..., C_M$, each containing $2^B$ vectors (for $B$-bit codes). A weight is encoded by choosing a single code from each codebook and summing them up. We denote this choice as a one-hot vector $b_m$, which results in the following representation for a group: $\sum_{m=1}^M  C_m b_{i j m}$.
This is similar to PTQ algorithms~\cite{frantar2022gptq}, except for using much more complex coding per group. 
To represent the full weights, we simply concatenate:

\vspace{-10px}\begin{equation}
\widehat{\mathbf{W}}_i {=} \sum_{m=1}^M  C_m b_{i, 1, m} \oplus ... \oplus \sum_{m=1}^M  C_m b_{i, d_{in}/g, m},
\label{eq:w_hat}
\end{equation}\vspace{-10px}

where $\oplus$ denotes concatenation and $b_{i j m} \in \mathbb{R}^{2^B}$ represents a one-hot code for the $i$-th output unit, $j$-th group of input dimensions and $m$-th codebook. 

\begin{figure}[h]
    \centering
    \vspace{0px}
    \includegraphics[width=0.49\textwidth]{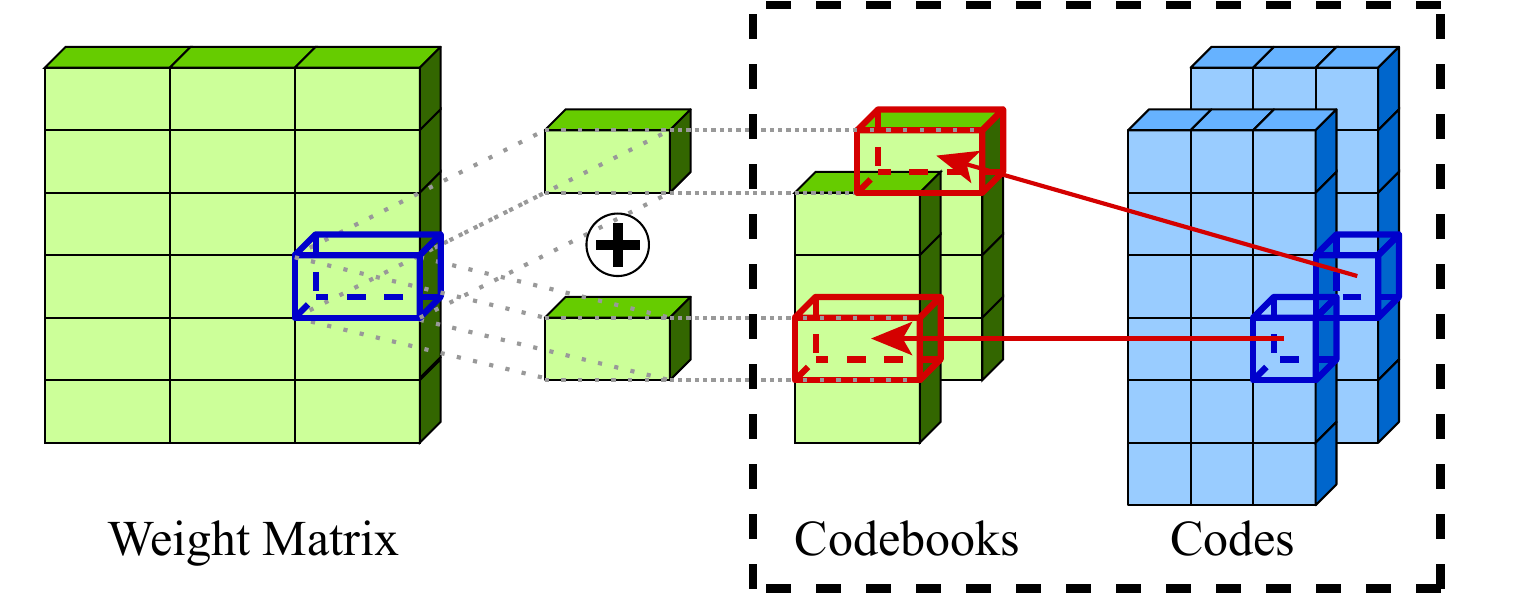}
    \vspace{-15px}
    \caption{Groups of weights are represented by a sum of codes selected from codebooks by corresponding indices.}\vspace{-5px}
    \label{fig:repr_scheme}
\end{figure}

Our algorithm will learn codebooks $C_m \in \mathbb{R}^{g \times 2^B}$ and the discrete codes represented by one-hot $b\in \mathbb{R}^{d_{out} \times d_{in}/g \times M \times 2^B}$. The resulting scheme encodes each group of $g$ weights using $M \cdot B$ bits and further requires  $g \cdot 2^B \cdot 16$ bits for FP16 codebooks. The  error becomes:

\vspace{-20px}\begin{equation}
    \underset{C, b}{\arg\min}||\mathbf{W} \mathbf{X} - \left(\text{Concat}_{i, j} \sum^M_{m=1} C_m b_{i, j, m}\right) \mathbf{X}||_2^2.
    \label{eq:main_obj}
\end{equation}\vspace{-10px}

To learn this weight representation, we initialize codebooks $C$ and codes $b$ by running residual K-means as in~\citet{rvq}. Specifically, the initialization algorithm proceeds as follows: first, it runs K-means clustering of weight groups and saves the resulting cluster indices. Next, it computes the quantization errors by subtracting the nearest cluster from every weight. Finally, the algorithm runs another round of K-means clustering, but this time on quantization errors instead of weights. Thus, each subsequent codebook is initialized to compensate the quantization error from previous codebooks. After initialization, we alter between updating codes $b_{i,j,m}$ and codebooks $C_m$ until the loss function~\eqref{eq:main_obj} stops improving up to the specified tolerance. Since codes are discrete and codebooks are continuous, and we are optimizing over multiple interacting layers, our approach has three phases, described in Algorithm~\ref{alg:highlvel} and detailed below.


\subsection{Phase 1: Beam search for codes}\label{sect:method_codes}

First, AQLM updates the codes $b_{i,j,m}$ to minimize the MSE objective~\eqref{eq:main_obj}. Similarly to~\citet{aq,lsq,lsq++}, we reformulate the objective in terms of a fully-connected discrete Markov Random Field (MRF) to take advantage of MRF solvers.

To simplify the derivation, let us first consider a special case of a single output unit ($d_{out}{=}1)$ and a single quantization group (i.e. $g{=}d_{in}$), to get rid of the concatenation operator: $||\mathbf{W} \mathbf{X} - \sum^M_{m=1} C_m b_m \mathbf{X}||_2^2$.
We rewrite this objective by expanding the squared difference:

\vspace{-20px}
\begin{multline}
||\mathbf{W} \mathbf{X} - \sum^M_{m=1} C_m b_m \mathbf{X}||_2^2 = ||\mathbf{W} \mathbf{X}||_2^2 - \\
    - 2 \left\langle \mathbf{W} \mathbf{X} \ \ ,\ \  \sum^M_{m=1} C_m b_m \mathbf{X} \right\rangle_F + ||\sum^M_{m=1} C_m b_m \mathbf{X}||_2^2 
\label{eq:difference_squared}
\end{multline}
\vspace{-10px}

Above, $\langle \cdot, \cdot \rangle_F$ denotes a Frobenius inner product of two matrices. 
Next, let us consider the three components of Eqn.~\eqref{eq:difference_squared} in isolation. First, note that $||\mathbf{W} \mathbf{X}||_2^2$ is constant in $b$ and can be ignored. The third component can be expanded further into pairwise dot products:

\vspace{-20px}\begin{equation}
\begin{aligned}
    ||\sum^M_{m=1} C_m b_m \mathbf{X}||_2^2 = \sum^M_{i=1} \sum^M_{j=1} \left\langle  C_i b_i \mathbf{X}, C_j b_j \mathbf{X} \right\rangle_F.
\end{aligned}
\end{equation}\vspace{-10px}

Note that both the second and third components rely on Frobenius products of $C_m b_m \mathbf{X}$-like matrices. These matrices can be inconvenient in practice: since $\mathbf{X} \in \mathbb{R}^{d_{in} \times n}$, the size of each matrix will scale with the size of calibration dataset $n$. To circumvent this, we rewrite the products as:

\vspace{-10px}\begin{equation}
\begin{aligned}
    \left\langle C_i b_i \mathbf{X} , C_j b_j \mathbf{X} \right\rangle_F = \left\langle C_i b_i \mathbf{X} \mathbf{X}^T , C_j b_j  \right\rangle_F.
\end{aligned}
\end{equation}

Thus one can pre-compute $\mathbf{X} \mathbf{X}^T \in \mathbb{R}^{d_{in} \times d_{in}}$. We will denote this type of product as $\left\langle \mathbf{A}, \mathbf{B} \right\rangle_{\mathbf{X} \mathbf{X}^T} \stackrel{\mathsf{def}}{=} \left\langle \mathbf{A} \mathbf{X} \mathbf{X}^T , \mathbf{B} \right\rangle_F$ in future derivations. 
Then, Eqn.~\eqref{eq:difference_squared} becomes:

\vspace{-15px}
\begin{multline}
    ||\mathbf{W} \mathbf{X} - \sum^M_{m=1} C_m b_m \mathbf{X}||_2^2 = ||\mathbf{W} \mathbf{X}||_2^2 - \\
    - 2 \sum^M_{m=1} \left\langle \mathbf{W}, C_m b_m\right\rangle_{\mathbf{XX}^T}
    + \sum^M_{i=1} \sum^M_{j=1} \left\langle C_i b_i, C_j b_j\right\rangle_{\mathbf{XX}^T} . 
    \label{eq:obj_mrf}
\end{multline}\vspace{-10px}

Finally, we generalize this equation to multiple output units ($d_{out}>1$) and quantization groups ($g{\ne}d_{in}$). For $d_{out}>1$, note that the original objective~\eqref{eq:main_obj} is additive with respect to output units: thus, we can apply \eqref{eq:obj_mrf} independently to each output dimension and sum up results. To support multiple input groups ($g{\ne}d_{in}$), we can treat each group as a separate codebook where only the codes for the active group are nonzero. Thus, we need to repeat each codebook $d_{in}/g$ times and pad it with zeros according to the active group.

It is now evident that minimizing~\eqref{eq:difference_squared} is equivalent to MAP inference in a Markov Random Field with $\left\langle \mathbf{W}, C_m b_m\right\rangle_{\mathbf{XX}^T}$ as unary potentials and $\left\langle C_i b_i, C_j b_j\right\rangle_{\mathbf{XX}^T}$ as pairwise potentials. While  finding the exact optimum is infeasible, 
prior work has shown that this type of MRF can be solved approximately via beam search or ICM~\cite{besag1986statistical}.

To solve this problem, we chose to adapt a beam search algorithm from~\citet{aq}. This algorithm maintains a beam of $k$ (beam size) best configurations for the codes, starting from the previous solution. On each step, the algorithm attempts to replace one code by trying all $2^B k$ alternatives and selecting the $k$ best based on MSE \eqref{eq:obj_mrf}.

Since the loss function is additive, changing one code only affects a small subset of loss components. Thus, we can compute the loss function efficiently by starting with a previous loss function (before code replacement), then adding and subtracting the components that changed during this iteration. These few loss components can be computed efficiently by multiplying with $\mathbf{X} \mathbf{X}^T$ ahead of beam search. The beam search runs over all $d_{out}$ output units in parallel. This is possible because encoding one output unit does not affect the objective~\eqref{eq:obj_mrf} of other units. 
Note that beam search is not necessarily the best  solution to this problem. AQ variants for retrieval~\cite{lsq,lsq++} use randomized ICM to find solutions faster. In this study, we chose beam search because it was easier to implement in ML frameworks like PyTorch/JAX.

\vspace{-5px}\subsection{Phase 2: Codebook update}\label{sect:method_codebook}

In the second phase, we find the optimal codebook vectors $C_1, ..., C_M$ that minimize the same squared error as the beam search. If we treat the codes $b$ as constants, minimizing \eqref{eq:main_obj} becomes a least squares problem for $C_m$. The original AQ algorithm solves this problem in closed form, relying on the fact that each vector dimension can be optimized independently. Our problem is complicated due to the presence of $\mathbf{X} \mathbf{X}^T$: the optimal value of one codebook coordinate depends on the values of all others. In principle, we could optimize $C_m$ in closed form, but it would require inverting a large matrix, or using iterative least squares solvers (e.g. conjugate gradients) specialized to this problem.

For simplicity, our current implementation defaults to using Adam~\cite{kingma2014adam} for approximately solving this minimization problem. In practice, this codebook tuning phase takes up a small fraction of the total compute time. We compute the objective as follows:

\vspace{-15px}
\begin{multline}
    ||\mathbf{W X} - \widehat{\mathbf{W}} \mathbf{X}||_2^2 = ||(\mathbf{W} - \widehat{\mathbf{W}}) \mathbf{X}||_2^2 = \\
    = \left\langle (\mathbf{W} - \widehat{\mathbf{W}}) \mathbf{X} \mathbf{X}^T, (\mathbf{W} - \widehat{\mathbf{W}})\right\rangle_F ,
\end{multline}\vspace{-10px}

 where $\widehat{\mathbf{W}}$ is the quantized weight matrix from~\ref{eq:w_hat}, and the $\mathbf{X} \mathbf{X}^T$ matrix is pre-computed. We optimize this objective by iterating (non-stochastic) full-batch gradient descent. 

For each update phase, our implementation runs 100 Adam steps with learning rate $10^{-4}$. However, we found that the final result is not sensitive to either of these parameters: training with smaller number of steps or learning rate achieves the same loss, but takes longer to converge. In future work, these hyperparameters could be eliminated by switching to dedicated least squares solver for codebooks.
Similarly to other algorithms, we also learn per-unit scales $s \in \mathbb{R}^{d_{out}}$ that are initialized as $s_i := ||\mathbf{W}_i||_2$ and updated alongside codebooks via the same optimizer (line 10 in Algorithm~\ref{alg:highlvel}). 

\subsection{Phase 3: Fine-tuning for intra-layer cohesion}\label{sect:method_finetuning}

So far, our algorithm compresses each weight matrix independently of the rest of the model. However, in practice, quantization errors interact differently between matrices. 
This issue is especially relevant in the case of extreme (2-bit) compression, where quantization errors are larger.

\begin{figure*}[!t]
    \centering
    \vspace{-5px}
    \includegraphics[scale=0.31]{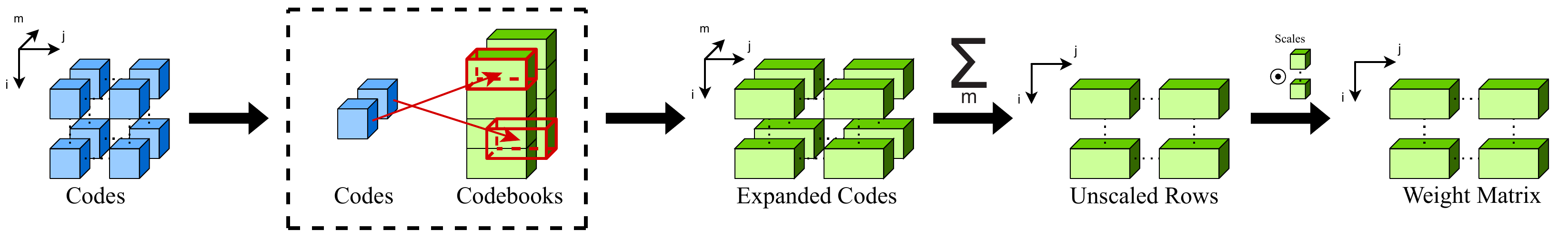}
    \vspace{-15px}
    \caption{AQLM compressed weight format. Horizontal and vertical axes are input features and output units, respectively. Depth  represents the codebook index. Reconstruction procedure, from left to right: i) compressed weight codes ii) zoom-in one weight group, each code is an index in its respective codebook iii) select codes from each codebook iv) add up codes as in \eqref{eq:w_hat} v) multiply by scales (one scale per output dimension).}
    \label{fig:dequant_scheme}
    \vspace{-10px}
\end{figure*}

Prior work addresses this issue via quantization-aware training (QAT), e.g.~\cite{gholami2021survey}. Instead of compressing the entire model in a single pass, they quantize model parameters gradually and train the remaining parameters to compensate for the quantization error. Unfortunately, running QAT in our setting is infeasible, since most modern LLMs are extremely expensive to train or even fine-tune. Thus, most PTQ algorithms for LLMs only adjust model parameters within the same linear layer~\cite{frantar2022gptq,lin2023awq,dettmers2023spqr}.

\setlength{\textfloatsep}{0.3cm}
\begin{algorithm}[t]
\caption{AQLM: Additive Quantization for LLMs}
\label{alg:highlvel}
\small
\begin{algorithmic}[1]
\vspace{-1px}\REQUIRE model, data

\STATE $\mathbf{X}_{block} := \texttt{model.input\_embeddings(data)}$
\FOR{$i = 1, \dots, \texttt{model.num\_layers} $}
    \STATE $\texttt{block := model.get\_block(i)}$
    \STATE $\mathbf{Y}_{block} := \texttt{block}(\mathbf{X}_{block})$
    \FOR{$\texttt{layer} \in \texttt{linear\_layers(block)}$}
        \STATE $\mathbf{W} := \texttt{layer.weight}$
        \STATE $\mathbf{X} := \texttt{layer\_inputs}(\texttt{layer}, \mathbf{X}_{block})$
        \STATE $C, b, s := \texttt{initialize}(\mathbf{W})$ \texttt{ // k-means}
        \WHILE{ \text{loss improves by at least $\tau$}}
            \STATE $C, s := \texttt{train\_Cs\_adam}(\mathbf{XX}^T, \mathbf{W}, C, b, s)$
            \STATE $b := \texttt{beam\_search}(\mathbf{XX}^T, \mathbf{W}, C, b, s)$
        \ENDWHILE
        \STATE \texttt{/* save for fine-tuning */}
        \STATE $\texttt{layer.weight} := \texttt{AQLMFormat}(C, b, s)$
    \ENDFOR
    \STATE $\theta := \texttt{trainable\_parameters}(\texttt{block})$
    \WHILE{ \text{loss improves by at least $\tau$}}
        \STATE $L := ||\texttt{block}(\mathbf{X}_{block}) - \mathbf{Y}_{block}||_2^2$
        \STATE $\theta := \texttt{adam}(\theta, \frac{\partial{L}}{\partial{\theta}})$
    \ENDWHILE
    \STATE $\mathbf{X}_{block} := \texttt{block}(\mathbf{X}_{block})$
\ENDFOR
\end{algorithmic}
\end{algorithm}

Here, we opt for a middle ground  by performing optimization at the level of individual transformer blocks, i.e. groups of 4-8 linear layers\footnote{This number depends on factors including the use of gated GLU activations, group query attention and QKV weight merging.} that constitute a single multi-head self-attention, followed by a single MLP layer\nocite{vaswani2017attention,shazeer2020glu}. Having quantized all linear layers within a single transformer block, we fine-tune its remaining parameters to better approximate the original outputs of that transformer block by backpropagating through the weight representation~\eqref{eq:w_hat}.

Concretely, we use the PyTorch\nocite{pytorch} autograd engine to differentiate the $||\text{block}(\mathbf{X}_{block}) - \mathbf{Y}_{block}||^2$ , where $\mathbf{X}_{block}$ are the inputs activations for that transformer block and $\mathbf{Y}_{block}$ are output activations of $\text{block}(\mathbf{X}_{block})$ recorded prior to quantization. We train the codebooks $C_m$, scale vectors $s$ and all non-quantized parameters (RMSNorm\nocite{rmsnorm} scales and biases), while keeping the codes $b_{i,j,m}$ frozen.
Similarly to Section~\ref{sect:method_codebook}, we train these parameters using Adam to minimize the MSE against the original block outputs (prior to quantization). This phase uses the same calibration data as for the individual layer quantization. The full procedure is summarized in Alg.~\ref{alg:highlvel}.

While fine-tuning blocks is more expensive than individual linear layers, it is still possible to quantize billion-parameter models on a single GPU in reasonable time. Also, since the algorithm only modifies a few trainable parameters, it uses little VRAM for optimizer states. This fine-tuning converges after a few iterations, as it starts from a good initial guess. In practice, fine-tuning transformer layers takes a minority (10-30\% or less) of the total calibration time. 

\begin{table*}[t!]
\vspace{-0.5em}
\footnotesize
\centering
\setlength\tabcolsep{2.37pt}
\renewcommand{\arraystretch}{1.15}
  \caption{Evaluation of quantized \textsc{Llama 2} models for \textbf{2-2.8 bits per parameter}, with an extra section for higher bitwidth. We report perplexity on WikiText-2~\citep{wikitext103} \& C4~\citep{C4} and accuracy for zero-shot tasks. The \textbf{Average accuracy} is the mean of 5 zero-shot tasks. Primary metrics are Wiki2 (PPL), C4 (PPL) and Average accuracy.}\label{tab:llama2bit}%
\begin{tabular}{lcc|cc|cccccc}
 \toprule
  \bf{Size} & \bf{Method} & \bf{Avg bits} & \bf{Wiki2$\downarrow$} & \bf{C4$\downarrow$} & \bf{WinoGrande$\uparrow$} & \bf{PiQA$\uparrow$} & \bf{HellaSwag$\uparrow$} & \bf{ArcE$\uparrow$} & \bf{ArcC$\uparrow$} & \bf{Average accuracy$\uparrow$}\\
  \midrule
  
  \multirow{5}{*}{7B} & -- & 16 & 5.12 & 6.63 & 67.25 & 78.45 & 56.69 & 69.32 & 40.02 & 62.35\\
  & AQLM & 2.02 & \textbf{6.59} & \textbf{8.54} & \textbf{65.67} & \textbf{74.76} & \textbf{49.55} & \textbf{63.68} & \textbf{32.76} & \textbf{57.28} \\
  & QuIP\# & 2.02 & 8.22 & 11.01 & 62.43 & 71.38 & 42.94 & 55.56 & 28.84 & 52.23 \\
  \cmidrule{2-11}
  & AQLM & 2.29 & 6.29 & 8.11 & 65.67 & 74.92 & 50.88 & 66.50 & 34.90 & 58.57 \\  
 \midrule

 \multirow{7}{*}{13B} & -- & 16 & 4.57 & 6.05 & 69.61 & 78.73 & 59.72 & 73.27 & 45.56 & 65.38\\
  & AQLM & 1.97 & \textbf{5.60} & \textbf{7.49} & \textbf{68.82} & \textbf{75.90} & \textbf{53.80} & \textbf{69.28} & \textbf{38.82} & \textbf{61.32} \\  
  & QuIP & 2.00 & 13.48 & 16.16 & 52.80 & 62.02 & 35.80 & 45.24 & 23.46 & 43.86 \\
  & QuIP\# & 2.01 & 6.06 & 8.07 & 63.38 & 74.76 & 51.58 & 64.06 & 33.96 & 57.55 \\
  \cmidrule{2-11}
  & AQLM & 2.19 & 5.37 & 7.16 & 67.64	& 77.37 & 55.03 & 70.29 & 38.65 & 61.80 \\ 
  & AQLM & 2.53 & 5.13 & 6.82 & 69.77	& 76.99 & 56.15 & 70.33 & 39.16 & 62.48 \\
  & AQLM & 2.76 & 4.94 & 6.54 & 68.98	& 77.58 & 57.71 & 72.90 & 43.60 & 64.15 \\
 \midrule

 \multirow{4}{*}{70B} & -- & 16 & 3.12 & 4.97 & 76.95 & 81.07 & 63.99 & 77.74 & 51.11 & 70.17 \\
  & AQLM & 2.07 & \bf{3.94} & \bf{5.72} & \bf{75.93} & \bf{80.43} & \bf{61.79} & \bf{77.68} & \bf{47.93} & \bf{68.75} \\
  & QuIP & 2.01 & 5.90 & 8.17 & 67.48 & 74.76 & 50.45 & 62.16 & 33.96 & 57.76 \\
  & QuIP\# & 2.01 & 4.16 & 6.01 & 74.11 & 79.76 & 60.01 & 76.85 & 47.61 & 67.67 \\
 \bottomrule
\end{tabular}

\end{table*}

\begin{table*}[t!]
\vspace{-0.5em}
\footnotesize
\centering
\setlength\tabcolsep{2.37pt}
\renewcommand{\arraystretch}{1.15}
  \caption{Evaluation of quantized \textsc{Llama 2} models for \textbf{3-3.1 bits per parameter}, with the same metrics as in Table~\ref{tab:llama2bit}.}\label{tab:llama3bit}%
\vspace{-5px}
\begin{tabular}{lcc|cc|cccccc}
 \toprule
  \bf{Size} & \bf{Method} & \bf{Avg bits} & \bf{Wiki2$\downarrow$} & \bf{C4$\downarrow$} & \bf{WinoGrande$\uparrow$} & \bf{PiQA$\uparrow$} & \bf{HellaSwag$\uparrow$} & \bf{ArcE$\uparrow$} & \bf{ArcC$\uparrow$} & \bf{Average accuracy$\uparrow$}\\
  \midrule
  
  \multirow{5}{*}{7B} & -- & 16 & 5.12 & 6.63 & 67.25 & 78.45 & 56.69 & 69.32 & 40.02 & 62.35\\
  & AQLM & 3.04 & \bf{5.46} & \bf{7.08} & \bf{66.93} & \bf{76.88} & \bf{54.12} & \bf{68.06} & \bf{38.40} & \bf{60.88} \\
  & GPTQ & 3.00 & 8.06 & 10.61 & 59.19 & 71.49 & 45.21 & 58.46 & 31.06 & 53.08 \\
  & SpQR & 2.98 & 6.20 & 8.20 & 63.54 & 74.81 & 51.85 & 67.42 & 37.71 & 59.07 \\
 \midrule

 \multirow{5}{*}{13B} & -- & 16 & 4.57 & 6.05 & 69.61 & 78.73 & 59.72 & 73.27 & 45.56 & 65.38\\
  & AQLM & 3.03 & \bf{4.82} & \bf{6.37} & 68.43 & \bf{77.26} & \bf{58.30} & \bf{70.88} & \bf{42.58} & \bf{64.49} \\
  & GPTQ & 3.00 & 5.85 & 7.86 & 63.93 & 76.50 & 53.47 & 65.66 & 38.48 & 59.61 \\
  & SpQR & 2.98 & 5.28 & 7.06 & 67.48 & 77.20 & 56.34 & 69.78 & 39.16 & 61.99 \\
  & QuIP & 3.00 & 5.12 & 6.79 & \bf{69.93} & 76.88 & 57.07 & 70.41 & 41.47 & 63.15 \\
 \midrule

 \multirow{5}{*}{70B} & -- & 16 & 3.12 & 4.97 & 76.95 & 81.07 & 63.99 & 77.74 & 51.11 & 70.17 \\
  & AQLM & 3.01 & \bf{3.36} & \bf{5.17} & \bf{77.19} & \bf{81.28} & \bf{63.23} & \bf{77.61} & \bf{50.00} & \bf{69.86} \\
  & GPTQ & 3.00 & 4.40 & 6.26 & 71.82 & 78.40 & 60.00 & 72.73 & 44.11 & 65.41 \\
  & SpQR & 2.98 & 3.85 & 5.63 & 74.66 & 80.52 & 61.95 & 75.93 & 48.04 & 68.22 \\
  & QuIP & 3.01 & 3.87 & 5.67 & 74.59 & 79.98 & 60.73 & 73.19 & 46.33 & 66.96 \\
 \bottomrule
\end{tabular}
\vspace{-10px}
\end{table*}

\vspace{-5px}\section{Experiments}

We evaluate the AQLM algorithm in typical scenarios for post-training quantization of modern LLMs. Our evaluation is focused on the \textsc{Llama 2} model family since it is a popular backbone for fine-tuned models or general LLM applications, e.g.~\cite{dettmers2023qlora}, and we also present results on Mistral-family models~\cite{jiang2024mixtral}.
In Section~\ref{sect:experiments_llama}, we evaluate the full AQ procedure for various \textsc{Llama 2} models and quantization bit-widths;  Section~\ref{sect:experiments_ablation} presents an ablation analysis for individual AQ components and implementation details.
\vspace{-5px}\subsection{Compression quality for modern LLMs}\label{sect:experiments_llama}

We report perplexity on WikiText-2~\cite{wikitext103} and C4~\cite{C4} validation sets. We also measure zero-shot accuracy on WinoGrande~\cite{DBLP:journals/cacm/winogrande2021}, PiQA~\cite{tata2003piqa}, HellaSwag~\cite{DBLP:conf/acl/hellaswag2019}, ARC-easy and ARC-challenge~\cite{arc_allenai} via the LM Eval Harness~\cite{eval-harness}. We follow the  evaluation setup of GPTQ~\cite{frantar2022gptq} and provide configurations for AQLM and baselines in Appendix~\ref{app:exp_config}.


We consider three main targets in terms of compression ranges: 2{-}2.8 bits, 3{-}3.1 bits, and 4{-}4.1 bits per model parameter. 
In the results below \emph{average bits per parameter} takes into account only quantized weights, we do not include
parameters kept in floating precision similarly to the related work. The details on the model size estimate
are provided in Appendix~\ref{app:model_size}. 
We compare AQ against GPTQ for 3\&4 bits~\cite{frantar2022gptq}, SpQR for 3\&4 bits~\cite{dettmers2023spqr}, QuIP in 2,3 \& 4 bits~\cite{chee2023quip} and QuIP\# for 2\&4 bits~\cite{quip-sharp}. While GPTQ and SpQR technically support 2-bit quantization, they perform poorly in  the 2-3 bit range. 
For QuIP, our adapted\footnote{The official QuIP (non-\#) code does not support \textsc{Llama 2}.} implementation shows acceptable performance for \textsc{Llama 2} 13B \& 70B but performs poorly for the 7B model.
We calibrate each algorithm using the subset of RedPajama dataset~\cite{together2023redpajama}, with a sequence length of 4096.

\begin{table*}[t!]
\footnotesize
\centering
\setlength\tabcolsep{2.37pt}
\renewcommand{\arraystretch}{1.15}
  \caption{Evaluation of quantized Mixtral~\cite{jiang2024mixtral} models for \textbf{2 bits}. The table reports perplexity on WikiText-2~\citep{wikitext103} and C4~\citep{C4}, as well as accuracy for zero-shot tasks. The \textbf{Average accuracy} column is the mean of 5 zero-shot task accuracies. Primary metrics are Wiki2 (PPL), C4 (PPL) and Average accuracy.}%
\label{tab:mixtral2}
\begin{tabular}{lcc|cc|cccccc}
 \toprule
  \bf{Size} & \bf{Method} & \bf{Avg bits} & \bf{Wiki2$\downarrow$} & \bf{C4$\downarrow$} & \bf{WinoGrande$\uparrow$} & \bf{PiQA$\uparrow$} & \bf{HellaSwag$\uparrow$} & \bf{ArcE$\uparrow$} & \bf{ArcC$\uparrow$} & \bf{Average accuracy$\uparrow$}\\
  \midrule
  
  \multirow{4}{*}{8x7B} & -- & 16 & 3.46 & 5.02 & 75.45 & 82.37 & 64.65 & 83.38 & 55.80 & 72.33\\
  & AQLM & 1.98 & \textbf{4.61} & \textbf{5.75} & \textbf{73.64} & \textbf{79.27} & 57.91 & \textbf{78.96} & \textbf{48.63} & \textbf{67.68} \\
  
  & QuIP\# &  2.01 & 4.75 & 5.89 & 71.11 & 79.05 & \textbf{58.23} & 77.57 & 45.73 & 66.34 \\

 \bottomrule
\end{tabular}

\end{table*}

The exact bit-widths for each method are dictated by parameters such as the number of codebooks and  code width. We report results for the $2{-}2.8$ and $3{-}3.1$ bitwidth ranges in Tables~\ref{tab:llama2bit} and \ref{tab:llama3bit}, respectively. 
Additional results for $4-4.1$ bits are deferred to Appendix~\ref{app:additional_experiments_Llama-2}.

The results show that AQLM outperforms the previous best PTQ algorithms across all settings, often by wide margins, especially at high compression. This holds both in terms of PPL across standard validation sets (Wiki-Text2 and C4), and accuracy across zero-shot tasks. Specifically, we observe the highest accuracy gains in the ``extreme'' 2-2.1 bits per parameter range, where the deviation from the uncompressed model becomes large for all methods. 

\paragraph{Mixtral quantization.} 
Table~\ref{tab:mixtral2} presents results on the Mixtral MoE, comparing against QuIP\# at 2-bits. (See Appendix \ref{app:additional_experiments_Mixtral} for full results.) 
AQLM outperforms QuIP\# in this case as well. Although the margins are lower compared to \textsc{Llama 2} models, they are still significant for ``harder'' tasks, such as Arc Challenge (+3 points).

\textbf{Pareto optimality of AQLM.} The significant error improvements raise the question of choosing the ``optimal'' model variant to maximize accuracy within a certain memory budget. 
For this, we follow \citet{dettmers2022case}: a quantized model is said to be Pareto-optimal if it maximizes accuracy at the same or lower total size (bytes). Despite rapid progress, prior art methods are \emph{not Pareto-optimal} at 2-bits: for instance, the previous best 2-bit \textsc{Llama 2} 13B (QuIP\#, Table~\ref{tab:llama2bit}) achieves Wiki2 PPL of 6.06, but one can get much lower 5.21 PPL by using a 7B model with 4-bit quantization, which is smaller (see Appendix Table~\ref{tab:llama4bit}).

AQLM compression to strictly 2 bits for the same model is also below Pareto-optimality, as it is outperformed by 4-bit AQLM compression for \textsc{Llama 2} 7B (5.21 vs 5.59). To find the Pareto-optimal quantization bitwidth, we run experiments between 2-3 bits per parameter and report them in Table~\ref{tab:llama2bit}, below horizontal bars. Thus,  the Pareto-optimal bitwidth for AQLM appears to be around 2.5 bits per parameter (Table~\ref{tab:llama2bit}), at which point we are comparable to 5-bit AQLM for \textsc{Llama 2} 7B (Appendix Table~\ref{tab:llama4bit}). In turn, the 2.76-bit AQLM on 13B outperforms  the \emph{uncompressed} 7B model. As such, AQLM is the first algorithm to achieve Pareto-optimality at less than 3 bits per parameter.

\begin{table*}[t!]
\vspace{-0.5em}
\footnotesize
\centering
\setlength\tabcolsep{2.37pt}
\renewcommand{\arraystretch}{1.15}
  \caption{Evaluation of quantized \textsc{Llama 2} with end-to-end fine-tuning, with the same metrics as in Table~\ref{tab:llama2bit}.}\label{tab:llama2bit_finetune}%
\begin{tabular}{lcc|cc|cccccc}
 \toprule
  \bf{Size} & \bf{Method} & \bf{Avg bits} & \bf{Wiki2$\downarrow$} & \bf{C4$\downarrow$} & \bf{WinoGrande$\uparrow$} & \bf{PiQA$\uparrow$} & \bf{HellaSwag$\uparrow$} & \bf{ArcE$\uparrow$} & \bf{ArcC$\uparrow$} & \bf{Average accuracy$\uparrow$}\\
  \midrule
  
  \multirow{5}{*}{7B} & -- & 16 & 5.12 & 6.63 & 67.25 & 78.45 & 56.69 & 69.32 & 40.02 & 62.35\\
  & AQLM$^{\bigstar}$ & 2.02 & \textbf{6.14} & \textbf{8.09} & \textbf{65.67} & \textbf{76.01} & 51.83 & 63.43 & 34.39 & 58.27 \\
  & QuIP\#$^{\bigstar}$ & 2.02 & 6.19 & 8.16 & 64.96 & 75.41 & \textbf{51.91} & \textbf{64.96} & \textbf{35.15} & \textbf{58.48} \\
  \cmidrule{2-11}
  & AQLM$^{\bigstar}$ & 2.29 & 5.92 & 7.86 & 63.77 & 76.93 & 52.82 & 66.16 & 36.95 & 59.33 \\  
 \midrule

 \multirow{7}{*}{13B} & -- & 16 & 4.57 & 6.05 & 69.61 & 78.73 & 59.72 & 73.27 & 45.56 & 65.38\\
 & AQLM$^{\bigstar}$ & 1.97 & \textbf{5.33} & \textbf{7.19} & \textbf{68.67} & 76.82 & \textbf{56.31} & \textbf{69.99} & \textbf{40.36} & \textbf{62.43} \\  
  & QuIP\#$^{\bigstar}$ & 2.01 & 5.35 & 7.20 & 67.64 & \textbf{77.26} & 56.04 & 69.02 & 39.85 & 61.96 \\
  \cmidrule{2-11}
  & AQLM$^{\bigstar}$ & 2.19 & 5.22 & 6.98 & 68.27 & 77.53 & 57.09 & 69.78 & 40.70 & 62.67 \\ 
 \midrule
 \multirow{4}{*}{70B} & -- & 16 & 3.12 & 4.97 & 76.95 & 81.07 & 63.99 & 77.74 & 51.11 & 70.17 \\
  & AQLM$^{\bigstar}$ & 2.07 & \bf{3.83} & \bf{5.62} & 74.35 & \bf{80.90} & 62.17 & 74.58 & \bf{48.98} & 68.20 \\
  & QuIP\#$^{\bigstar}$  & 2.01 & 3.91 & 5.71 & \bf{74.66} & 79.54 & \bf{62.52} & \bf{77.06} & 47.61 & \bf{68.28} \\
 \bottomrule
\end{tabular}

\end{table*}

\subsection{End-to-end fine-tuning experiments}\label{sect:fullfinetune}


Subsequent work in QuIP\#~\cite{quip-sharp} improves upon our block-wise protocol (Section~\ref{sect:method_finetuning}) by fine-tuning the entire model to mimimize KL divergence. Here, we analyze how this end-to-end fine-tuning translates to AQLM. We follow the setup from QuIP\#~\cite{quip-sharp} and run end-to-end fine-tuning with default parameters (see Appendix~\ref{app:end-to-end_finetuning}). Table~\ref{tab:llama2bit_finetune} reports our results for 2-bit quantization using AQLM and QuIP\# with end-to-end fine-tuning.
We report additional results in this setup in Tables~\ref{tab:llama3bit_finetune},~\ref{tab:mistral234}~and~\ref{tab:mmlu_gsm8k_evaluation} in supplementary materials. To differentiate between two versions, we mark quantized models with end-to-end fine-tuning with ${\bigstar}$. 
Overall, end-to-end fine-tuning improves both QuIP\# and AQLM, reaching comparable accuracy for both methods. Additionally, we notice that the boost from end-to-end fine-tuning is more profound on 2-bit quantized models with diminishing returns for 3 bits and above. Finally, we can see that 2.19-bit AQLM with end-to end fine-tuning on 13B is comparable with an \emph{uncompressed} 7B model achieving Pareto optimality on zero-shot tasks.

\vspace{-1px}
\subsection{Ablation analysis}\label{sect:experiments_ablation}
In Appendix~\ref{app:ablations}, we examine key design choices regarding initialization, alternating optimization, the impact of the fine-tuning, and sensitivity to hyperparameters. In brief, we first find that the \emph{residual K-means initialization} is critical for \emph{fast} algorithm convergence: when compared with random initialization, it needs significantly fewer training iterations. We also compare different hyperparameter configurations for the same bitwidth, varying the number of codebooks and group size.
Second, to validate our calibration fine-tuning procedure, we compare it against 1) no fine-tuning, 2) fine-tuning only of non-linear layers (e.g. RMSNorm) but not of codebook parameters, and 3) fine-tuning only the codebooks (but not other layers). 
The results, presented in full in Appendix~\ref{app:ablations}, show that  fine-tuning the \emph{codebook parameters} has the highest impact on accuracy, by far, while fine-tuning the RMSNorm only has minor impact. This validates our choice of leveraging the calibration set for learned codebooks. 

Further, we observe that, increasing the number of sample sequences in the range 128 to 4096 leads to a gradual PPL improvement, but with diminishing returns. This is true for both initial AQLM calibraton and fine-tuning. In this respect, AQLM  benefits more from larger calibration sets (similarly to QuIP\#), as opposed to direct methods like GPTQ which saturate accuracy at around 256 input sequences. 
 Finally, we investigate various options for investing a given bit budget, comparing e.g. longer codes (e.g. 1x15) vs multiple codebooks with shorter codes (e.g. 2x8). 

\subsection{Inference Speed}\label{sect:experiments_inference}

Although our primary objective is to maximize accuracy for a given model size, AQLM can also be practical in terms of inference latency. To demonstrate this, we implemented efficient GPU and CPU kernels for a few hardware-friendly configurations of AQLM. 
The results can be found in  Table~\ref{tab:inference_speed}.
For GPU inference, we targeted quantized \textsc{Llama 2} models with 16-bit codebooks, corresponding to 2.07 bits for \textsc{Llama 2} 70B, 2.19 bits for 13B, and 2.29 bits for 7B models (see Table~\ref{tab:llama2bit}, \ref{tab:llama2bit_finetune}), as well as a 2x8-bit codebook model with perplexity 6.57 on Wiki2(see Table~\ref{tab:2x8codebooks}).
For each model we benchmark the matrix-vector multiplication subroutine performance on a standard layer. 
The results show that AQLM can execute at speeds comparable to or better than  FP16. 
End-to-end generative numbers with HuggingFace integration can be found in Appendix~\ref{app:generation}: for instance, we can achieve ${\approx}14$ tokens/s on \textsc{Llama 2} 70B in this setting.
We observe that multiple smaller codebooks allow efficient  GPU cache utilization, leading to  greater speedup, at the price of slightly lower accuracy.

\begin{table}[h!]
    \vspace{-3px}
    \caption{Speed of the FP16 gate\_proj layer matrix-vector multiplication in PyTorch, and relative AQLM speedups. } 
    \label{tab:inference_speed}
    \vspace{-10px}
    \begin{center}
        \begin{tabular}{lccc}
            \toprule
            \bf{Llama 2} & \bf{7B} & \bf{13B} & \bf{70B} \\
            \midrule
            \multicolumn{4}{c}{2 bit speedup over FP16 on Nvidia RTX 3090 GPU} \\
            \midrule
            Original (float16)                    &  129 $\mu$s   &  190 $\mu$s     & 578 $\mu$s     \\
            AQLM (Table~\ref{tab:llama2bit})      &  x1.31    &  x1.20      & x1.20      \\
            AQLM ($2\times$8-bit)                 &  x1.57    &  x1.82      & x3.05      \\
            \midrule
            \multicolumn{4}{c}{2 bit speedup over FP32 on Intel i9 CPU, 8 cores} \\
            \midrule
            Original (float32)                    &  1.83 ms    &  3.12 ms     & 11.31 ms     \\
            AQLM ($2\times$8-bit)                 &  x2.75    &  x3.54      & x3.69      \\
            AQLM ($4\times$8-bit)                 &  x2.55    &  x3.02      & x4.07     \\
            AQLM ($8\times$8-bit)                 &  x2.29    &  x2.68      & x4.03      \\

            \bottomrule
        \end{tabular}
    \end{center}
    \vspace{-10px}
\end{table}

Next, we explore how to leverage AQLM to accelerate CPU inference. As discussed in Section~\ref{sect:related_NNS}, additive quantization can compute dot products efficiently if the codebook size is small. One way to achieve it for AQLM is to replace each 16-bit codebook with a number of smaller 8-bit ones. This leads to higher quantization error, but still outperforms the baselines in terms of accuracy (see Appendix Table~\ref{tab:groupsxcodebooks}). 
The results in Table~\ref{tab:inference_speed} show that this also allows for up to 4x faster inference relative to FP32 on CPU.

\vspace{-0.5em}
\section{Conclusion and Future Work}



We presented AQLM, a new form of additive quantization (AQ) targeting LLM compression, which significantly improved the  
state-of-the-art results for LLM quantization in the regime of 2 and 3 bits per weight. 
In terms of limitations, AQLM is more computationally-expensive than direct post-training quantization methods, such as RTN or GPTQ, specifically because of the use of a more complex coding representation. 
Yet, despite the more sophisticated encoding and decoding, we have shown AQLM lends itself to efficient implementation on both CPU and GPU.
Overall, we find it remarkable that, using AQLM, massive LLMs can be executed accurately and efficiently using little memory. 

While AQLM already achieves substantial improvements in low-bit quantization, there are several promising directions for further improvement that we did not explore in this work. One such direction is better fine-tuning strategies. In Section~\ref{sect:fullfinetune} we found that better fine-tuning algorithms~\cite{quip-sharp,malinovskii2024pv} can significantly improve quantized model accuracy. We believe that AQLM can benefit from a more systematic exploration of fine-tuning algorithms in future work.
Another promising direction is generalizing AQLM to other quantization scenarios. While our work is focused around LLM quantization, the underlying algorithm can potentially be adapted to other problems, e.g. quantizing computer vision models, compressing LLM attention caches for long sequences, and others.




\section*{Acknowledgements}
Authors would like to thank Ruslan Svirschevski for his help in solving technical issues with AQLM and baselines. We also thank Tim Dettmers for helpful discussions on the structure of weights in modern LLMs and size-accuracy trade-offs. The authors would also like to thank Daniil Pavlov for his assistance with CPU benchmarking. The authors would also like to thank contributors and community from Github repository\footnote{\url{https://github.com/Vahe1994/AQLM/}} for helping to improve the code and the text of the paper.  Finally, authors would like to thank the communities of ML enthusiasts known as LocalLLaMA\footnote{\url{https://www.reddit.com/r/LocalLLaMA/}} and Petals community on discord\footnote{\url{https://github.com/bigscience-workshop/petals/}} for the crowd wisdom about running LLMs on consumer devices.
Egiazarian Vage and Denis Kuznedelev and Andrei Panferov were supported by the grant for research centers in the field of AI provided by the Analytical Center for the Government of the Russian Federation (ACRF) in accordance with the agreement on the provision of subsidies (identifier of the agreement 000000D730321P5Q0002) and the agreement with HSE University No. 70-2021-00139

\section*{Impact Statement}
This paper presents work whose goal is to advance the field of 
Machine Learning. There are many potential societal consequences 
of our work, none which we feel must be specifically highlighted here.

\bibliographystyle{icml2024}\bibliography{bibliography}

\begin{thebibliography}{59}
\providecommand{\natexlab}[1]{#1}
\providecommand{\url}[1]{\texttt{#1}}
\expandafter\ifx\csname urlstyle\endcsname\relax
  \providecommand{\doi}[1]{doi: #1}\else
  \providecommand{\doi}{doi: \begingroup \urlstyle{rm}\Url}\fi

\bibitem[Babenko \& Lempitsky(2014)Babenko and Lempitsky]{aq}
Babenko, A. and Lempitsky, V.
\newblock Additive quantization for extreme vector compression.
\newblock In \emph{Proceedings of the IEEE Conference on Computer Vision and Pattern Recognition}, pp.\  931--938, 2014.

\bibitem[Besag(1986)]{besag1986statistical}
Besag, J.
\newblock On the statistical analysis of dirty pictures.
\newblock \emph{Journal of the Royal Statistical Society Series B: Statistical Methodology}, 48\penalty0 (3):\penalty0 259--279, 1986.

\bibitem[Biderman et~al.(2023)Biderman, Schoelkopf, Anthony, Bradley, O'Brien, Hallahan, Khan, Purohit, Prashanth, Raff, et~al.]{biderman2023pythia}
Biderman, S., Schoelkopf, H., Anthony, Q., Bradley, H., O'Brien, K., Hallahan, E., Khan, M.~A., Purohit, S., Prashanth, U.~S., Raff, E., et~al.
\newblock Pythia: A suite for analyzing large language models across training and scaling.
\newblock \emph{arXiv preprint arXiv:2304.01373}, 2023.

\bibitem[Blalock \& Guttag(2021)Blalock and Guttag]{maddness2021}
Blalock, D. and Guttag, J.
\newblock Multiplying matrices without multiplying.
\newblock In \emph{International Conference on Machine Learning}, pp.\  992--1004. PMLR, 2021.

\bibitem[Burton et~al.(1983)Burton, Shore, and Buck]{vq1}
Burton, D., Shore, J., and Buck, J.
\newblock A generalization of isolated word recognition using vector quantization.
\newblock In \emph{ICASSP '83. IEEE International Conference on Acoustics, Speech, and Signal Processing}, volume~8, pp.\  1021--1024, 1983.
\newblock \doi{10.1109/ICASSP.1983.1171915}.

\bibitem[Chee et~al.(2023)Chee, Cai, Kuleshov, and Sa]{chee2023quip}
Chee, J., Cai, Y., Kuleshov, V., and Sa, C.~D.
\newblock Quip: 2-bit quantization of large language models with guarantees, 2023.

\bibitem[Chen et~al.(2019)Chen, Wang, and Pan]{Chen_Wang_Pan_2019}
Chen, S., Wang, W., and Pan, S.~J.
\newblock Deep neural network quantization via layer-wise optimization using limited training data.
\newblock \emph{Proceedings of the AAAI Conference on Artificial Intelligence}, 33\penalty0 (01):\penalty0 3329--3336, Jul. 2019.
\newblock \doi{10.1609/aaai.v33i01.33013329}.
\newblock URL \url{https://ojs.aaai.org/index.php/AAAI/article/view/4206}.

\bibitem[Chen et~al.(2010)Chen, Guan, and Wang]{rvq}
Chen, Y., Guan, T., and Wang, C.
\newblock Approximate nearest neighbor search by residual vector quantization.
\newblock \emph{Sensors}, 10\penalty0 (12):\penalty0 11259--11273, 2010.

\bibitem[Clark et~al.(2018)Clark, Cowhey, Etzioni, Khot, Sabharwal, Schoenick, and Tafjord]{arc_allenai}
Clark, P., Cowhey, I., Etzioni, O., Khot, T., Sabharwal, A., Schoenick, C., and Tafjord, O.
\newblock Think you have solved question answering? try arc, the ai2 reasoning challenge.
\newblock \emph{arXiv preprint arXiv:1803.05457}, 2018.

\bibitem[Cobbe et~al.(2021)Cobbe, Kosaraju, Bavarian, Chen, Jun, Kaiser, Plappert, Tworek, Hilton, Nakano, Hesse, and Schulman]{gsm8k}
Cobbe, K., Kosaraju, V., Bavarian, M., Chen, M., Jun, H., Kaiser, L., Plappert, M., Tworek, J., Hilton, J., Nakano, R., Hesse, C., and Schulman, J.
\newblock Training verifiers to solve math word problems.
\newblock \emph{CoRR}, abs/2110.14168, 2021.
\newblock URL \url{https://arxiv.org/abs/2110.14168}.

\bibitem[Computer(2023)]{together2023redpajama}
Computer, T.
\newblock Redpajama: an open dataset for training large language models, 2023.
\newblock URL \url{https://github.com/togethercomputer/RedPajama-Data}.

\bibitem[Dettmers \& Zettlemoyer(2022)Dettmers and Zettlemoyer]{dettmers2022case}
Dettmers, T. and Zettlemoyer, L.
\newblock The case for 4-bit precision: k-bit inference scaling laws.
\newblock \emph{arXiv preprint arXiv:2212.09720}, 2022.

\bibitem[Dettmers et~al.(2022)Dettmers, Lewis, Belkada, and Zettlemoyer]{dettmers2022llm}
Dettmers, T., Lewis, M., Belkada, Y., and Zettlemoyer, L.
\newblock {LLM}.int8(): 8-bit matrix multiplication for transformers at scale.
\newblock \emph{Advances in Neural Information Processing Systems 35: Annual Conference on Neural Information Processing Systems 2022, NeurIPS 2022}, 2022.

\bibitem[Dettmers et~al.(2023{\natexlab{a}})Dettmers, Pagnoni, Holtzman, and Zettlemoyer]{dettmers2023qlora}
Dettmers, T., Pagnoni, A., Holtzman, A., and Zettlemoyer, L.
\newblock {QLoRA}: Efficient finetuning of quantized llms.
\newblock \emph{arXiv preprint arXiv:2305.14314}, 2023{\natexlab{a}}.

\bibitem[Dettmers et~al.(2023{\natexlab{b}})Dettmers, Svirschevski, Egiazarian, Kuznedelev, Frantar, Ashkboos, Borzunov, Hoefler, and Alistarh]{dettmers2023spqr}
Dettmers, T., Svirschevski, R., Egiazarian, V., Kuznedelev, D., Frantar, E., Ashkboos, S., Borzunov, A., Hoefler, T., and Alistarh, D.
\newblock Spqr: A sparse-quantized representation for near-lossless llm weight compression.
\newblock \emph{arXiv preprint arXiv:2306.03078}, 2023{\natexlab{b}}.

\bibitem[Fern{\'a}ndez-Marqu{\'e}s et~al.(2023)Fern{\'a}ndez-Marqu{\'e}s, AbouElhamayed, Lane, and Abdelfattah]{FernndezMarqus2023AreWT}
Fern{\'a}ndez-Marqu{\'e}s, J., AbouElhamayed, A.~F., Lane, N.~D., and Abdelfattah, M.~S.
\newblock Are we there yet? product quantization and its hardware acceleration.
\newblock \emph{ArXiv}, abs/2305.18334, 2023.
\newblock URL \url{https://api.semanticscholar.org/CorpusID:258967539}.

\bibitem[Frantar \& Alistarh(2023)Frantar and Alistarh]{frantar2023qmoe}
Frantar, E. and Alistarh, D.
\newblock Qmoe: Practical sub-1-bit compression of trillion-parameter models.
\newblock \emph{arXiv preprint arXiv:2310.16795}, 2023.

\bibitem[Frantar et~al.(2022{\natexlab{a}})Frantar, Ashkboos, Hoefler, and Alistarh]{frantar2022gptq}
Frantar, E., Ashkboos, S., Hoefler, T., and Alistarh, D.
\newblock Gptq: Accurate post-training quantization for generative pre-trained transformers.
\newblock \emph{arXiv preprint arXiv:2210.17323}, 2022{\natexlab{a}}.

\bibitem[Frantar et~al.(2022{\natexlab{b}})Frantar, Singh, and Alistarh]{frantar2022obc}
Frantar, E., Singh, S.~P., and Alistarh, D.
\newblock {Optimal Brain Compression}: A framework for accurate post-training quantization and pruning.
\newblock \emph{arXiv preprint arXiv:2208.11580}, 2022{\natexlab{b}}.
\newblock Accepted to NeurIPS 2022, to appear.

\bibitem[Gao et~al.(2021)Gao, Tow, Biderman, Black, DiPofi, Foster, Golding, Hsu, McDonell, Muennighoff, Phang, Reynolds, Tang, Thite, Wang, Wang, and Zou]{eval-harness}
Gao, L., Tow, J., Biderman, S., Black, S., DiPofi, A., Foster, C., Golding, L., Hsu, J., McDonell, K., Muennighoff, N., Phang, J., Reynolds, L., Tang, E., Thite, A., Wang, B., Wang, K., and Zou, A.
\newblock A framework for few-shot language model evaluation, September 2021.
\newblock URL \url{https://doi.org/10.5281/zenodo.5371628}.

\bibitem[Ge et~al.(2013)Ge, He, Ke, and Sun]{opq}
Ge, T., He, K., Ke, Q., and Sun, J.
\newblock Optimized product quantization.
\newblock \emph{IEEE transactions on pattern analysis and machine intelligence}, 36\penalty0 (4):\penalty0 744--755, 2013.

\bibitem[Gholami et~al.(2021)Gholami, Kim, Dong, Yao, Mahoney, and Keutzer]{gholami2021survey}
Gholami, A., Kim, S., Dong, Z., Yao, Z., Mahoney, M.~W., and Keutzer, K.
\newblock A survey of quantization methods for efficient neural network inference.
\newblock \emph{arXiv preprint arXiv:2103.13630}, 2021.

\bibitem[Gray(1984)]{vq2}
Gray, R.
\newblock Vector quantization.
\newblock \emph{IEEE ASSP Magazine}, 1\penalty0 (2):\penalty0 4--29, 1984.
\newblock \doi{10.1109/MASSP.1984.1162229}.

\bibitem[Guo et~al.(2016)Guo, Kumar, Choromanski, and Simcha]{guo2016quantization}
Guo, R., Kumar, S., Choromanski, K., and Simcha, D.
\newblock Quantization based fast inner product search.
\newblock In \emph{Artificial intelligence and statistics}, pp.\  482--490. PMLR, 2016.

\bibitem[Hendrycks et~al.(2020)Hendrycks, Burns, Basart, Zou, Mazeika, Song, and Steinhardt]{mmlu}
Hendrycks, D., Burns, C., Basart, S., Zou, A., Mazeika, M., Song, D., and Steinhardt, J.
\newblock Measuring massive multitask language understanding.
\newblock \emph{CoRR}, abs/2009.03300, 2020.
\newblock URL \url{https://arxiv.org/abs/2009.03300}.

\bibitem[Hinton et~al.(2015)Hinton, Vinyals, and Dean]{hinton2015distilling}
Hinton, G., Vinyals, O., and Dean, J.
\newblock Distilling the knowledge in a neural network, 2015.

\bibitem[Jegou et~al.(2010)Jegou, Douze, and Schmid]{pq}
Jegou, H., Douze, M., and Schmid, C.
\newblock Product quantization for nearest neighbor search.
\newblock \emph{IEEE transactions on pattern analysis and machine intelligence}, 33\penalty0 (1):\penalty0 117--128, 2010.

\bibitem[Jiang et~al.(2023)Jiang, Sablayrolles, Mensch, Bamford, Chaplot, Casas, Bressand, Lengyel, Lample, Saulnier, et~al.]{jiang2023mistral}
Jiang, A.~Q., Sablayrolles, A., Mensch, A., Bamford, C., Chaplot, D.~S., Casas, D. d.~l., Bressand, F., Lengyel, G., Lample, G., Saulnier, L., et~al.
\newblock Mistral 7b.
\newblock \emph{arXiv preprint arXiv:2310.06825}, 2023.

\bibitem[Jiang et~al.(2024)Jiang, Sablayrolles, Roux, Mensch, Savary, Bamford, Chaplot, Casas, Hanna, Bressand, et~al.]{jiang2024mixtral}
Jiang, A.~Q., Sablayrolles, A., Roux, A., Mensch, A., Savary, B., Bamford, C., Chaplot, D.~S., Casas, D. d.~l., Hanna, E.~B., Bressand, F., et~al.
\newblock Mixtral of experts.
\newblock \emph{arXiv preprint arXiv:2401.04088}, 2024.

\bibitem[Kim et~al.(2023)Kim, Hooper, Gholami, Dong, Li, Shen, Mahoney, and Keutzer]{kim2023squeezellm}
Kim, S., Hooper, C., Gholami, A., Dong, Z., Li, X., Shen, S., Mahoney, M.~W., and Keutzer, K.
\newblock Squeezellm: Dense-and-sparse quantization.
\newblock \emph{arXiv preprint arXiv:2306.07629}, 2023.

\bibitem[Kingma \& Ba(2015)Kingma and Ba]{kingma2014adam}
Kingma, D.~P. and Ba, J.
\newblock Adam: A method for stochastic optimization.
\newblock \emph{International Conference on Learning Representations (ICLR)}, 2015.

\bibitem[Li et~al.(2017)Li, Ni, Zhang, Yang, and Gao]{li2017performance}
Li, Z., Ni, B., Zhang, W., Yang, X., and Gao, W.
\newblock Performance guaranteed network acceleration via high-order residual quantization, 2017.

\bibitem[Lin et~al.(2023)Lin, Tang, Tang, Yang, Dang, and Han]{lin2023awq}
Lin, J., Tang, J., Tang, H., Yang, S., Dang, X., and Han, S.
\newblock Awq: Activation-aware weight quantization for llm compression and acceleration.
\newblock \emph{arXiv preprint arXiv:2306.00978}, 2023.

\bibitem[Malinovskii et~al.(2024)Malinovskii, Mazur, Ilin, Kuznedelev, Burlachenko, Yi, Alistarh, and Richtarik]{malinovskii2024pv}
Malinovskii, V., Mazur, D., Ilin, I., Kuznedelev, D., Burlachenko, K., Yi, K., Alistarh, D., and Richtarik, P.
\newblock Pv-tuning: Beyond straight-through estimation for extreme llm compression.
\newblock \emph{arXiv preprint arXiv:2405.14852}, 2024.

\bibitem[Martinez et~al.(2016)Martinez, Clement, Hoos, and Little]{lsq}
Martinez, J., Clement, J., Hoos, H.~H., and Little, J.~J.
\newblock Revisiting additive quantization.
\newblock In \emph{Computer Vision--ECCV 2016: 14th European Conference, Amsterdam, The Netherlands, October 11-14, 2016, Proceedings, Part II 14}, pp.\  137--153. Springer, 2016.

\bibitem[Martinez et~al.(2018)Martinez, Zakhmi, Hoos, and Little]{lsq++}
Martinez, J., Zakhmi, S., Hoos, H.~H., and Little, J.~J.
\newblock Lsq++: Lower running time and higher recall in multi-codebook quantization.
\newblock In \emph{Proceedings of the European Conference on Computer Vision (ECCV)}, pp.\  491--506, 2018.

\bibitem[McCarter \& Dronen(2022)McCarter and Dronen]{McCarter2022LookupsAN}
McCarter, C. and Dronen, N.
\newblock Look-ups are not (yet) all you need for deep learning inference.
\newblock \emph{ArXiv}, abs/2207.05808, 2022.
\newblock URL \url{https://api.semanticscholar.org/CorpusID:250491319}.

\bibitem[Merity et~al.(2016)Merity, Xiong, Bradbury, and Socher]{wikitext103}
Merity, S., Xiong, C., Bradbury, J., and Socher, R.
\newblock Pointer sentinel mixture models.
\newblock \emph{arXiv preprint arXiv:1609.07843}, 2016.

\bibitem[Nagel et~al.(2020)Nagel, Amjad, Van~Baalen, Louizos, and Blankevoort]{nagel2020up}
Nagel, M., Amjad, R.~A., Van~Baalen, M., Louizos, C., and Blankevoort, T.
\newblock Up or down? {A}daptive rounding for post-training quantization.
\newblock In \emph{International Conference on Machine Learning (ICML)}, 2020.

\bibitem[Norouzi \& Fleet(2013)Norouzi and Fleet]{norouzi13}
Norouzi, M. and Fleet, D.~J.
\newblock Cartesian k-means.
\newblock In \emph{Proceedings of the IEEE Conference on computer Vision and Pattern Recognition}, pp.\  3017--3024, 2013.

\bibitem[Ozan et~al.(2016)Ozan, Kiranyaz, and Gabbouj]{competitveq}
Ozan, E.~C., Kiranyaz, S., and Gabbouj, M.
\newblock Competitive quantization for approximate nearest neighbor search.
\newblock \emph{IEEE Transactions on Knowledge and Data Engineering}, 28\penalty0 (11):\penalty0 2884--2894, 2016.
\newblock \doi{10.1109/TKDE.2016.2597834}.

\bibitem[Park et~al.(2022)Park, Park, Kwon, Kim, Lee, and Lee]{park2022nuqmm}
Park, G., Park, B., Kwon, S.~J., Kim, B., Lee, Y., and Lee, D.
\newblock {nuQmm}: Quantized matmul for efficient inference of large-scale generative language models.
\newblock \emph{arXiv preprint arXiv:2206.09557}, 2022.

\bibitem[Paszke et~al.(2019)Paszke, Gross, Massa, Lerer, Bradbury, Chanan, Killeen, Lin, Gimelshein, Antiga, Desmaison, Kopf, Yang, DeVito, Raison, Tejani, Chilamkurthy, Steiner, Fang, Bai, and Chintala]{pytorch}
Paszke, A., Gross, S., Massa, F., Lerer, A., Bradbury, J., Chanan, G., Killeen, T., Lin, Z., Gimelshein, N., Antiga, L., Desmaison, A., Kopf, A., Yang, E., DeVito, Z., Raison, M., Tejani, A., Chilamkurthy, S., Steiner, B., Fang, L., Bai, J., and Chintala, S.
\newblock {PyTorch}: An imperative style, high-performance deep learning library.
\newblock In \emph{Conference on Neural Information Processing Systems (NeurIPS)}. 2019.

\bibitem[Raffel et~al.(2020)Raffel, Shazeer, Roberts, Lee, Narang, Matena, Zhou, Li, and Liu]{C4}
Raffel, C., Shazeer, N., Roberts, A., Lee, K., Narang, S., Matena, M., Zhou, Y., Li, W., and Liu, P.
\newblock Exploring the limits of transfer learning with a unified text-to-text transformer.
\newblock \emph{Journal of Machine Learning Research}, 21\penalty0 (140):\penalty0 1--67, 2020.

\bibitem[Sakaguchi et~al.(2021)Sakaguchi, Bras, Bhagavatula, and Choi]{DBLP:journals/cacm/winogrande2021}
Sakaguchi, K., Bras, R.~L., Bhagavatula, C., and Choi, Y.
\newblock Winogrande: an adversarial winograd schema challenge at scale.
\newblock \emph{Commun. {ACM}}, 64\penalty0 (9):\penalty0 99--106, 2021.
\newblock \doi{10.1145/3474381}.
\newblock URL \url{https://doi.org/10.1145/3474381}.

\bibitem[Scao et~al.(2022)Scao, Fan, Akiki, Pavlick, Ili{\'c}, Hesslow, Castagn{\'e}, Luccioni, Yvon, Gall{\'e}, et~al.]{scao2022bloom}
Scao, T.~L., Fan, A., Akiki, C., Pavlick, E., Ili{\'c}, S., Hesslow, D., Castagn{\'e}, R., Luccioni, A.~S., Yvon, F., Gall{\'e}, M., et~al.
\newblock Bloom: A 176b-parameter open-access multilingual language model.
\newblock \emph{arXiv preprint arXiv:2211.05100}, 2022.

\bibitem[Shazeer(2020)]{shazeer2020glu}
Shazeer, N.
\newblock Glu variants improve transformer, 2020.

\bibitem[Tata \& Patel(2003)Tata and Patel]{tata2003piqa}
Tata, S. and Patel, J.~M.
\newblock {PiQA}: An algebra for querying protein data sets.
\newblock In \emph{International Conference on Scientific and Statistical Database Management}, 2003.

\bibitem[{TII UAE}(2023)]{falcon2023}
{TII UAE}.
\newblock The {Falcon} family of large language models.
\newblock \url{https://huggingface.co/tiiuae/falcon-40b}, May 2023.

\bibitem[Touvron et~al.(2023)Touvron, Lavril, Izacard, Martinet, Lachaux, Lacroix, Rozi{\`e}re, Goyal, Hambro, Azhar, et~al.]{touvron2023llama}
Touvron, H., Lavril, T., Izacard, G., Martinet, X., Lachaux, M.-A., Lacroix, T., Rozi{\`e}re, B., Goyal, N., Hambro, E., Azhar, F., et~al.
\newblock Llama: Open and efficient foundation language models.
\newblock \emph{arXiv preprint arXiv:2302.13971}, 2023.

\bibitem[Tseng et~al.(2024)Tseng, Chee, Sun, Kuleshov, and Sa]{quip-sharp}
Tseng, A., Chee, J., Sun, Q., Kuleshov, V., and Sa, C.~D.
\newblock Quip\#: Even better llm quantization with hadamard incoherence and lattice codebooks, 2024.

\bibitem[Vaswani et~al.(2017)Vaswani, Shazeer, Parmar, Uszkoreit, Jones, Gomez, Kaiser, and Polosukhin]{vaswani2017attention}
Vaswani, A., Shazeer, N., Parmar, N., Uszkoreit, J., Jones, L., Gomez, A.~N., Kaiser, L., and Polosukhin, I.
\newblock Attention is all you need.
\newblock \emph{arXiv preprint arXiv:1706.03762}, 2017.

\bibitem[Xiao et~al.(2022)Xiao, Lin, Seznec, Demouth, and Han]{xiao2022smoothquant}
Xiao, G., Lin, J., Seznec, M., Demouth, J., and Han, S.
\newblock Smoothquant: Accurate and efficient post-training quantization for large language models.
\newblock \emph{arXiv preprint arXiv:2211.10438}, 2022.

\bibitem[Yao et~al.(2022)Yao, Aminabadi, Zhang, Wu, Li, and He]{yao2022zeroquant}
Yao, Z., Aminabadi, R.~Y., Zhang, M., Wu, X., Li, C., and He, Y.
\newblock Zeroquant: Efficient and affordable post-training quantization for large-scale transformers.
\newblock \emph{arXiv preprint arXiv:2206.01861}, 2022.

\bibitem[Zellers et~al.(2019)Zellers, Holtzman, Bisk, Farhadi, and Choi]{DBLP:conf/acl/hellaswag2019}
Zellers, R., Holtzman, A., Bisk, Y., Farhadi, A., and Choi, Y.
\newblock Hellaswag: Can a machine really finish your sentence?
\newblock In Korhonen, A., Traum, D.~R., and M{\`{a}}rquez, L. (eds.), \emph{Proceedings of the 57th Conference of the Association for Computational Linguistics, {ACL} 2019, Florence, Italy, July 28- August 2, 2019, Volume 1: Long Papers}, pp.\  4791--4800. Association for Computational Linguistics, 2019.
\newblock \doi{10.18653/v1/p19-1472}.
\newblock URL \url{https://doi.org/10.18653/v1/p19-1472}.

\bibitem[Zhang \& Sennrich(2019)Zhang and Sennrich]{rmsnorm}
Zhang, B. and Sennrich, R.
\newblock Root mean square layer normalization.
\newblock \emph{CoRR}, abs/1910.07467, 2019.
\newblock URL \url{http://arxiv.org/abs/1910.07467}.

\bibitem[Zhang et~al.(2022)Zhang, Roller, Goyal, Artetxe, Chen, Chen, Dewan, Diab, Li, Lin, et~al.]{zhang2022opt}
Zhang, S., Roller, S., Goyal, N., Artetxe, M., Chen, M., Chen, S., Dewan, C., Diab, M., Li, X., Lin, X.~V., et~al.
\newblock Opt: Open pre-trained transformer language models.
\newblock \emph{arXiv preprint arXiv:2205.01068}, 2022.

\bibitem[Zhang et~al.(2014)Zhang, Du, and Wang]{compositeq}
Zhang, T., Du, C., and Wang, J.
\newblock Composite quantization for approximate nearest neighbor search.
\newblock In \emph{International Conference on Machine Learning}, pp.\  838--846. PMLR, 2014.

\bibitem[Zhou et~al.(2017)Zhou, Wang, Wen, He, and Zou]{Zhou2017}
Zhou, S.-C., Wang, Y.-Z., Wen, H., He, Q.-Y., and Zou, Y.-H.
\newblock Balanced quantization: An effective and efficient approach to quantized neural networks.
\newblock \emph{Journal of Computer Science and Technology}, 32\penalty0 (4):\penalty0 667--682, Jul 2017.
\newblock ISSN 1860-4749.
\newblock \doi{10.1007/s11390-017-1750-y}.
\newblock URL \url{https://doi.org/10.1007/s11390-017-1750-y}.

\end{thebibliography}

\appendix

\newpage
\appendix
\onecolumn


\section{End-to-end fine-tuning}
 \label{app:end-to-end_finetuning}
The block-wise finetuning procedure, introduced in \ref{sect:method_finetuning}, 
considerably improves performance of compressed models. 
However, block-wise finetuning optimizes the loss only at the level of a current transformer block
and is agnostic of the actual task of interest. To minimize the target loss, one can run backpropagation through the whole model
and directly optimize all trainable parameters to minimize a model-level objective function. 

This allows to search for globally optimal parameters, as opposed to sequentially selected ones, during block-wise finetuning.

One can minimize the error between the quantized model and the floating-point model
on some calibration set. The parameters being optimized (namely the codebooks, scales and the non-quantized parameters) typically constitute a small fraction of the total number of parameters in the original model. Therefore, the proposed distillation method resembles parameter-efficient finetuning (PEFT) in both optimization and memory footprint. 

To transfer the knowledge from the original model to the quantized one, we adopt Knowledge Distillation \cite{hinton2015distilling} 
where the student model is taught to mimic the output of a teacher given the same input. We follow the setup from QuIP\# \cite{quip-sharp} that uses KL divergence between the outputs of teacher and
student models:
\begin{equation}
\mathcal{L} = \frac{1}{N} \sum_{i=0}^{N-1} D_{KL} (p_s (\mathbf{x}_i), p_t (\mathbf{x}_i))
\end{equation}

Above $D_{KL}$ is the Kullback–Leibler divergence and $p_s$, $p_t$ are the student and teacher probabilities given input sequence $\mathbf{x}_i$. 

Despite its simplicity, this fine-tuning procedure often significantly improves performance of the compressed model. 

We fine-tune all models on $4{-}16$M training tokens: $1{-}4$k sequences of length 4k
for \textsc{Llama 2} models \cite{touvron2023llama} and 512 sequences of length 8k for Mixtral \cite{jiang2024mixtral}. We fine-tune on the same data as during initial calibration (i.e. samples from RedPajama \cite{together2023redpajama}) and use Adam~\cite{kingma2014adam} optimizer with constant learning rate $10^{-5}$ without weight decay. Batch size is set to $8{-}16$ sequences. 
A single epoch of fine-tuning turns out to be sufficient, and longer training leads to marginal improvements. 
 
\section{Code reproducibility}
We share the code for our method in the GitHub repository \url{https://github.com/Vahe1994/AQLM/tree/AQLM_camera_ready}. The hyperparameters for our experimental setup are discussed in  Appendix \ref{app:exp_config}.

\section{Experimental Configurations}
\label{app:exp_config}
\textbf{Hardware.}
In all of our experiments, we used either Nvidia A100 or H100. The number of GPUs varied from 1 to 8. We used activation offloading to lower pick memory usage. To evaluate inference speed on GPU we used consumer-grade GPU Nvidia 3090 and for CPU setup we used Intel core i9 13900k. 

\textbf{Calibration set.} All methods were calibrated on a slice of RedPajama-v1 dataset \cite{together2023redpajama} for both \textsc{Llama} and Mistral/Mixtral family models.
We used the same context length as models were trained on, for \textsc{Llama 2} 4096 and for Mistral/Mixtral 8192.

For \textsc{Llama 2} experiments, we used 8M tokens as a calibration set for SpQR, GPTQ, and AQLM. Quip, however, was calibrated on 4M tokens due to OOM errors when trying to use more samples. Taking into account the fact that after 2M tokens improvement of methods results is fairly small we chose to report these numbers as is. 
For Quip\#, we used \textsc{Llama 2} and Mistral's quantized models provided by authors in their GitHub repository. To the best of our knowledge, they used 6k samples for calibration with a context length of 4096/8192.\\
For Mixtral we calibrated both our method and QUIP\# on 8M tokens with context length 8192.

\textbf{Hyperparameters.} 

For \textbf{GPTQ} for both 3 and 4 bits we used a standard set of parameters without grouping and with permutation order act\_order.

\textbf{SpQR} method was evaluated with base 2 and 3 bit-width with group size of 16 and 3 bits for zeros and scales. Outliers rate was chosen such that average bit will be close to 3 and 4 bits respectively.

\textbf{Quip} was adapted to work on the \textsc{Llama} family and was calibrated with 1024 samples and 4096 context length.

\textbf{Quip\#} For \textsc{Llama 2} and Mistral models we used the officially published quantized models. For Mixtral we adapted the code to work with the model's architecture and quantized it with the recommended set of parameters. For both AQLM and QuIP\#, we don't quantize gate linear layer in Mixtral, because it contains relatively small   amount of paramters and have severe impact on performance.

\textbf{AQLM} For to get 2, 3, 4 bits: we used 1 codebook size of $2^{15}$ or  $2^{16}$, with groups of 8 for 2 bits. For 3 bits we used 2 codebooks size of $2^{12}$ with groups of 8. Finally for 4 bits we used  2 codebooks size of $2^{15}$ or $2^{16}$ with groups of 8. 

Both for finetuning~\ref{sect:method_finetuning} and  codebooks update~\ref{sect:method_codebook} we used Adam optimizer \cite{kingma2014adam} with learning rate of $10^{-4}$, $\beta_1=0.90$ and $\beta_2=0.95$.
We used early stopping both for the finetuning phase and for the codebook optimization phase, by stopping when the least square error not decreasing more than some threshold. In our experiments the  threshold varies between  $10^{-2}$ and $10^{-3}$.

Hyperparameters for end-end fine-tuning discussed at the end of Appendix~\ref{app:end-to-end_finetuning}.

\section{Quantization time}

AQLM quantization takes considerably longer to calibrate than simpler quantization methods such as RTN or GPTQ. This only impacts quantization time, not inference time.

Quantizing a 7B model with default configuration takes about 1 day on a single A100 gpu. Similarly, quantizing a 70B model on a single GPU would take 10-14 days. However, the procedure can be parallelized across multiple GPU: 7B quantization takes ~14h on 2 GPUs, and 70B quantization takes 3-4 days on 8 GPUs. 

Full model fine-tuning with default configuration for 7B model would take 3-6 hours on four A100 , for 13B 10-16 hours on four A100, and for 70B 1-2 days on 8 A100.

Finally, the quantization time is dependent on the quantization configuration and its hyperparameters. Tweaking these parameters, e.g. by reducing the number of beams, can achieve notable speedups of 2-4x during quantization, but at the cost of lower model accuracy.

\begin{table*}[t!]
\vspace{-0.5em}
\footnotesize
\centering
\setlength\tabcolsep{2.37pt}

\renewcommand{\arraystretch}{1.15}
  \caption{Evaluation of quantized \textsc{Llama 2} end-to-end fine-tuned models for \textbf{3-3.1 bits per parameter}, with the same metrics as in Table~\ref{tab:llama2bit}.}\label{tab:llama3bit_finetune}%
\vspace{-5px}
\begin{tabular}{lcc|cc|cccccc}
 \toprule
  \bf{Size} & \bf{Method} & \bf{Avg bits} & \bf{Wiki2$\downarrow$} & \bf{C4$\downarrow$} & \bf{WinoGrande$\uparrow$} & \bf{PiQA$\uparrow$} & \bf{HellaSwag$\uparrow$} & \bf{ArcE$\uparrow$} & \bf{ArcC$\uparrow$} & \bf{Average accuracy$\uparrow$}\\
  \midrule
  
  \multirow{5}{*}{7B} & -- & 16 & 5.12 & 6.63 & 67.25 & 78.45 & 56.69 & 69.32 & 40.02 & 62.35\\
  & AQLM$^{\bigstar}$ & 3.04 & \bf{5.38} & \bf{7.01} & 65.35 & \bf{77.31} & \bf{55.49} & 66.79 & 38.48 & 60.68 \\
  & QuIP\#$^{\bigstar}$ & 3.04 & 5.41 & 7.04 & \bf{66.85} & \bf{77.31} & 55.32 & \textbf{68.43} & \textbf{38.99} & \textbf{61.38} \\
 \midrule

 \multirow{5}{*}{13B} & -- & 16 & 4.57 & 6.05 & 69.61 & 78.73 & 59.72 & 73.27 & 45.56 & 65.38\\
  & AQLM$^{\bigstar}$ & 3.03 & \bf{4.78} & \bf{6.33} & \bf{68.75} & \bf{78.45} & \bf{58.54} & \bf{72.94} & \bf{42.75} & \bf{64.29} \\
  & QuIP\#$^{\bigstar}$ & 3.01 & \bf{4.78}	 & 6.35 & 68.03 & 77.86 & 57.56 & 72.18 & 41.38 & 63.40 \\
 \midrule
 \multirow{5}{*}{70B} & -- & 16 & 3.12 & 4.97 & 76.95 & 81.07 & 63.99 & 77.74 & 51.11 & 70.17 \\
  & AQLM$^{\bigstar}$ & 3.01 & 3.36 & 5.17 & 75.30 & 80.69 & 63.48 & \bf{77.99} & 50.26 & 69.54 \\
  & QuIP\#$^{\bigstar}$ & 3.00 & \bf{3.35} & \bf{5.15} & \bf{76.40} & \bf{81.45} & \textbf{63.53} & 77.53 & \textbf{50.77} & \textbf{69.94} \\
 \bottomrule
\end{tabular}
\vspace{-10px}
\end{table*}
\section{Ablation analysis}\label{app:ablations}
The AQLM algorithm makes several design choices that need to be validated separately: initialization,  alternating optimization, the fine-tuning protocol, and the choice of hyperparameters. Here, we study how each of these components affect results.

\begin{figure}[h]
    \centering
    \vspace{-5px}
    \includegraphics[width=0.99\linewidth]{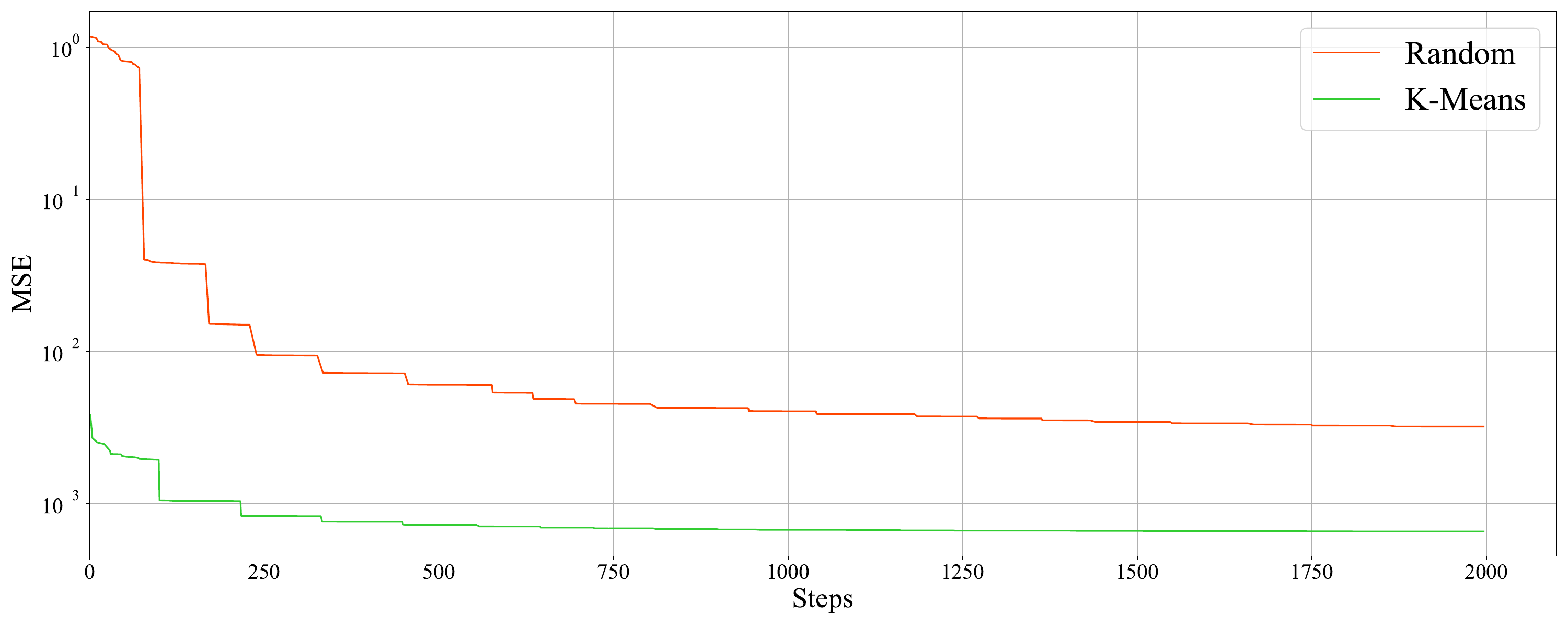}
    \vspace{-15px}
    \caption{MSE loss learning curves of AQLM trained on the self attention q\_proj linear layer of 10-th block in the \textsc{Llama 2} 70B model.}
    \vspace{-5px}
    \label{fig:kmeansvsrandom}
\end{figure}

\textbf{Initialization.} As discussed in Section~\ref{sect:method}, we initialize AQLM with residual K-means to obtain a good initial guess for both codes and codebooks. That is, we run K-means for the weight matrix, then subtract the nearest cluster from each weight, and run K-means again M times. A simple baseline would be to initialize all codes uniformly at random. We compare the two initialization strategies for the problem of quantizing a single linear layer within \textsc{Llama 2} 70B model to 3 bits per parameter. We quantize groups of 8 consecutive weights using 2 codebooks, 12 bit each. Each codebook contains ${2^{12}}$ learnable values. As we can see in Figure~\ref{fig:kmeansvsrandom}, AQLM with K-means initialization needs significantly fewer training iterations to achieve the desired loss. The difference is so drastic that we expect that running AQLM with a random initialization would require extremely high runtimes to accurately quantize the largest models.

\textbf{Fine-tuning.} Next, we validate the fine-tuning procedure. We compare the full block fine-tuning (default) against three alternatives: i) no fine-tuning at all, ii) fine-tuning only non-linear layers (i.e. RMSNorm), but not the AQ parameters, and iii) fine-tuning only the AQ parameters, but not the non-linear layers. Table~\ref{tab:finetune} summarizes our results: fine-tuning the entire model or only AQ parameters achieves competitive performance, while training only RMSNorm scales is comparable to not fine-tuning at all. We attribute these observations to the fact that over 99\% of quantized layer parameters are contained in AQ codebooks $C_m$, whereas the remaining parameters are small 1-dimensional tensors. This validates the use of the AQ approach, as many competing algorithms do not have learnable per-layer codebooks. Notably, QuIP\# uses a shared fixed lattice instead. We also note that, even without fine-tuning, AQLM is competitive to previous state-of-the-art results.

\begin{table}[ht!]
    \caption{Ablation analysis of AQLM with different fine-tuning restrictions on Llama-2 7B model at 2.02 bit width.}
    \label{tab:finetune}
    \begin{center}
        \begin{tabular}{lcc}
            \toprule
            \bf{Name} & \bf{Wiki2$\downarrow$} & \bf{C4$\downarrow$}\\
            \midrule
            w/o & 8.18 &  10.59\\
            RMSnorm & 8.31 & 10.46\\
            AQ params & 6.92 & 8.85\\
            Full & 6.93 & 8.84\\
            
            \bottomrule
        \end{tabular}
    \end{center}
    \vspace{-5px}
\end{table}

\textbf{Number of samples.} We verify our choice of calibration hyperparameters. Traditionally, most PTQ algorithms use several hundred calibration sequences (e.g. \citet{frantar2022gptq} has 128). In our experiments, we evaluate both AQLM and baselines with additional calibration data. Our original motivation for that was to avoid potential overfitting when fine-tuning entire transformer blocks.
To test this assumption, we run our algorithm with different calibration set sizes, varying from 128 to 4096 sequences. For each size, we report the average perplexity on WikiText-2 over 3 runs, along with standard deviations.
The results in Table~\ref{tab:nsample_abl} demonstrate that increasing the number of samples leads to gradual reduction in perplexity with seemingly diminishing returns.
Since the perplexity is still monotonically improving from 128 to 4096 samples, it is possible that larger sample sizes would yield further improvements.

\textbf{Number of codebooks vs groups.}
Finally, we conducted an additional set of experiments on \textsc{LLama 2} 7B models to see perplexity  dependence on simultaneous change on WikiText-2 of both codebooks and groups keeping  compression rate fixed to 2bits. We present both AQLM with and without end-to-end fine-tuning in  Table~\ref{tab:groupsxcodebooks}.


\noindent \begin{minipage}[t]{0.50\textwidth}
    
    \captionof{table}{WikiText-2 PPL as a function of calibration set size for Llama 2 (7B) quantized to 2.3 bits with AQLM, averaged over 3 runs. SD stands for adjusted standard deviation.} 
    \label{tab:nsample_abl}
    \vspace{-10px}
    \begin{center}
        \begin{tabular}{lcc}
            \toprule
            \bf{\# of samples} & \bf{Average PPL} & \bf{SD}\\
            \midrule
            128 & 6.994 &  0.127\\
            256 & 6.584 & 0.031\\
            512 & 6.455 & 0.005\\
            1024 & 6.353 & 0.008\\
            2048 & 6.297 & 0.018\\
            4096 & 6.267 & 0.005\\
            \bottomrule
        \end{tabular}
    \end{center}

\end{minipage}
\hspace{0.05\textwidth}
\begin{minipage}[t]{0.45\textwidth}
    
    \captionof{table}{WikiText-2 PPL as a function of from  groups and number of codebook for Llama 2 (7B) quantized with approximately 2 bits quantization.} 
    \label{tab:groupsxcodebooks}
    \vspace{-10px}
    \begin{center}
        \begin{tabular}{lcc}
            \toprule
            \bf{Method} & \bf{Setup} & \bf{Average PPL}\\
            \midrule
            \multirow{4}{*}{AQLM} & 2x8gs8 & 7.6107\\
            & 4x8gs16 & 8.1394\\
             & 8x8gs32 & 7.3755\\
             & 15x8gs64 & 7.8459\\
            \midrule
            \multirow{3}{*}{AQLM$^{\bigstar}$} & 2x8gs8 & 6.5746\\
             & 8x8gs32 & 6.6126\\
             & 15x8gs64 & 6.6602\\
            \bottomrule
        \end{tabular}
    \end{center}
\end{minipage}
\vspace{1cm}

\section{Additional experiments}
\label{app:additional_experiments}
In this section we report additional experimental results for Mixtral\cite{jiang2024mixtral}, Mistral7B\cite{jiang2023mistral} and \textsc{LLama 2} model.

\begin{table*}[t!]
\vspace{-0.4em}
\footnotesize
\centering
\setlength\tabcolsep{2.37pt}
\renewcommand{\arraystretch}{1.15}
  \caption{Evaluation of quantized \textsc{Llama 2} models for \textbf{4+ bits per parameter}. The table reports perplexity on WikiText-2~\citep{wikitext103} and C4~\citep{C4}, as well as accuracy for zero-shot tasks. The \textbf{Average accuracy} column is the mean of 5 zero-shot task accuracies. Primary metrics are Wiki2 (PPL), C4 (PPL) and Average accuracy.}\label{tab:llama4bit}%
\vspace{-5px}
\begin{tabular}{lcc|cc|cccccc}
 \toprule
  \bf{Size} & \bf{Method} & \bf{Avg bits} & \bf{Wiki2$\downarrow$} & \bf{C4$\downarrow$} & \bf{WinoGrande$\uparrow$} & \bf{PiQA$\uparrow$} & \bf{HellaSwag$\uparrow$} & \bf{ArcE$\uparrow$} & \bf{ArcC$\uparrow$} & \bf{Average accuracy$\uparrow$}\\
  \midrule
  
  \multirow{6}{*}{7B} & -- & 16 & 5.12 & 6.63 & 67.25 & 78.45 & 56.69 & 69.32 & 40.02 & 62.35\\
  & AQLM & 4.04 & \bf{5.21} & \bf{6.75} & 67.32 & 78.24 & \bf{55.99} & \bf{70.16} & \bf{41.04} & \bf{62.55} \\
  & GPTQ & 4.00 & 5.49 & 7.20 & \textbf{68.19} & 76.61 & 55.44 & 66.20 & 36.77 & 60.64 \\
  & SpQR & 3.98 & 5.28 & 6.87 & 66.93 & \textbf{78.35} & 56.10 & 69.11 & 39.68 & 62.17 \\
  & QuIP\# & 4.02 & 5.29 & 6.86 & 66.85 & 77.91 & 55.78 & 68.06 & 39.68 & 61.66 \\
  \cmidrule{2-11}
  & AQLM & 5.02 & 5.16 & 6.68 & 67.40 & 78.29 & 56.53 & 68.94 & 39.93 & 62.22 \\
 \midrule

 \multirow{6}{*}{13B} & -- & 16 & 4.57 & 6.05 & 69.61 & 78.73 & 59.72 & 73.27 & 45.56 & 65.38\\
  & AQLM & 3.94 & \bf{4.65} & \bf{6.14} & 69.85 & 78.35 & \bf{59.27} & 73.32 & 44.80 & 65.12 \\
  & GPTQ & 4 & 4.78 & 6.34 & \textbf{70.01} & 77.75 & 58.67 & 70.45 & 42.49 & 63.87 \\
  & SpQR & 3.98 & 4.69 & 6.20 & 69.69 & 78.45 & 59.25 & 71.21 & 44.52 & 64.42 \\
  & QuIP & 4.00 & 4.76 & 6.29 & 69.69 & \bf{79.00} & 58.91 & 73.27 & \bf{44.88} & \bf{65.15} \\
  & QuIP\# & 4.01 & 4.68 & 6.20 & 69.38 & 77.91 & 58.86 & \bf{73.74} & 44.63 & 64.90 \\
 \midrule

 \multirow{6}{*}{70B} & -- & 16 & 3.12 & 4.97 & 76.95 & 81.07 & 63.99 & 77.74 & 51.11 & 70.17\\
  & AQLM & 4.14 & \bf{3.19} & \bf{5.03} & 76.48 & \textbf{81.50} & 63.69 & 77.31 & 50.68 & 69.93 \\
  & GPTQ & 4.00 & 3.35 & 5.15 & 75.61 & 81.23 & 63.47 & 76.81 & 49.15 & 69.25 \\
  & SpQR & 3.97 & 3.25 & 5.07 & 76.01 & 81.28 & \textbf{63.71} & 77.36 & 49.15 & 69.50 \\
  & QuIP & 4.00 & 3.58 & 5.38 & 76.01 & 80.25 & 61.97 & 74.28 & 47.01 & 67.90 \\
  & QuIP\# & 4.01 & 3.22 & 5.05 & \textbf{76.80} & 81.45 & 63.51 & \bf{78.37} & \textbf{50.85} & \bf{70.20} \\
\cmidrule{2-11}
 & AQLM & 3.82 & 3.21 & \bf{5.03} & 76.32 & 80.90 & 63.69 & 77.61 & 50.34 & 69.77\\
 \bottomrule
\end{tabular}
\end{table*}

\subsection{Mixtral}\label{app:additional_experiments_Mixtral}
We report the results for Mixtral\cite{jiang2024mixtral} MoE-type model for 3 and 4 bits in Table~\ref{tab:mixtral34}. In the 4 bit case, performance of QuIP\# and AQLM are very similar across all metrics and close to uncompressed FP16 model.

\begin{table*}[t!]
\vspace{-0.5em}
\footnotesize
\centering
\setlength\tabcolsep{2.37pt}
\renewcommand{\arraystretch}{1.15}
  \caption{Evaluation of quantized Mixtral~\cite{jiang2024mixtral} models for \textbf{3 and 4 bits per parameter}. The table reports perplexity on WikiText-2~\citep{wikitext103} and C4~\citep{C4}, as well as accuracy for zero-shot tasks. The \textbf{Average accuracy} column is the mean of 5 zero-shot task accuracies. The primary metrics are Wiki2 (PPL, lower is better), C4 (PPL, lower is better) and Average accuracy (percentage, higher is better).}%
\label{tab:mixtral34}
\begin{tabular}{lcc|cc|cccccc}
 \toprule
  \bf{Size} & \bf{Method} & \bf{Avg bits} & \bf{Wiki2$\downarrow$} & \bf{C4$\downarrow$} & \bf{WinoGrande$\uparrow$} & \bf{PiQA$\uparrow$} & \bf{HellaSwag$\uparrow$} & \bf{ArcE$\uparrow$} & \bf{ArcC$\uparrow$} & \bf{Average accuracy$\uparrow$}\\
  \midrule
  
  \multirow{2}{*}{3-bit} & -- & 16.00 & 3.46 & 5.02 & 75.45 & 82.37 & 64.65 & 83.38 & 55.80 & 72.33\\
  & AQLM & 3.02 & \textbf{3.79} & \textbf{5.17} & \textbf{75.45} & \textbf{81.61} & \textbf{63.25} & \textbf{81.90} & \textbf{53.92} & \textbf{71.23} \\
 \midrule
  \multirow{3}{*}{4-bit} & -- & 16.00 & 3.46 & 5.02 & 75.45 & 82.37 & 64.65 & 83.38 & 55.80 & 72.33\\
  & AQLM & 3.915 & \textbf{3.57} & \textbf{5.07} & 74.82 & \textbf{81.99} & \textbf{64.23} & \textbf{83.12} & 54.61 & 71.75 \\
  & QuIP\# &  4.000 & 3.60 & 5.08 & \textbf{76.56} & \textbf{81.99} & 63.92 & 82.62 & \textbf{54.78} & \textbf{71.97} \\

 \bottomrule
\end{tabular}

\end{table*}
\subsection{\textsc{Llama 2}}
\label{app:additional_experiments_Llama-2}
We show results for 4 bit quantization of the \textsc{Llama 2} models in Table~\ref{tab:llama4bit}. We can see that AQLM  outperforms other methods in terms of perplexity and has the best or close to the best results. We also report results of perplexity for our quantized 2x8 codebooks models in Table~\ref{tab:2x8codebooks}.


\begin{table*}[t!]
\vspace{-0.5em}
\footnotesize
\centering
\setlength\tabcolsep{2.37pt}
\renewcommand{\arraystretch}{1.15}
  \caption{Evaluation of quantized \textsc{Llama 2} for \textbf{2x8groupsize8} codebooks models. We report perplexity on WikiText-2~\citep{wikitext103} \& C4~\citep{C4} and accuracy for zero-shot tasks. The \textbf{Average accuracy} is the mean of 5 zero-shot tasks. Primary metrics are Wiki2 (PPL), C4 (PPL) and Average accuracy.}\label{tab:2x8codebooks}%
\begin{tabular}{lcc|cc|cccccc}
 \toprule
  \bf{Size} & \bf{Method} & \bf{Avg bits} & \bf{Wiki2$\downarrow$} & \bf{C4$\downarrow$} & \bf{WinoGrande$\uparrow$} & \bf{PiQA$\uparrow$} & \bf{HellaSwag$\uparrow$} & \bf{ArcE$\uparrow$} & \bf{ArcC$\uparrow$} & \bf{Average accuracy$\uparrow$}\\
  \midrule
  
  \multirow{3}{*}{7B} & -- & 16 & 5.12 & 6.63 & 67.25 & 78.40 & 56.67 & 69.36 & 39.51 & 62.24\\
  & AQLM & 2 & 7.61 & 9.68 & 62.27 & 71.87 & 46.41 & 61.03 & 30.03 & 54.32 \\
  \cmidrule{2-11}
  & AQLM$^{\bigstar}$ & 2 & 6.57 & 8.60 & 63.22 & 74.54 & 50.08 & 61.28 & 31.83 & 56.19 \\  
 \midrule

 \multirow{3}{*}{13B} & -- & 16 & 4.57 & 6.05 & 69.61 & 78.73 & 59.72 & 73.27 & 45.56 & 65.38\\
 & AQLM & 2 & 6.54 & 8.77 & 55.96 & 71.06 & 48.29 & 62.50 & 31.40 & 53.84 \\  
  \cmidrule{2-11}
  & AQLM$^{\bigstar}$ & 2 & 5.63 & 7.55 & 6385 & 77.04 & 54.19 & 67.85 & 37.20 & 60.03 \\ 
 \midrule
 \multirow{2}{*}{70B} & -- & 16 & 3.12 & 4.97 & 76.95 & 81.07 & 63.99 & 77.74 & 51.11 & 70.17 \\
  & AQLM$^{\bigstar}$ & 2 & 4.21 & 5.99 & 73.48 & 79.54 & 61.29 & 74.49 & 46.84 & 67.13 \\
 \bottomrule
\end{tabular}

\end{table*}

\begin{table*}[t!]
\vspace{-0.5em}
\footnotesize
\centering
\setlength\tabcolsep{2.37pt}
\renewcommand{\arraystretch}{1.15}
  \caption{Evaluation of quantized Mistral7B~\cite{jiang2023mistral} models for \textbf{2, 3 and 4 bits per parameter}: perplexity on WikiText-2~\citep{wikitext103} and C4~\citep{C4}, as well as accuracy for zero-shot tasks. The \textbf{Average accuracy} column is the mean of 5 zero-shot task accuracies. Primary metrics are Wiki2 (PPL), C4 (PPL) and Average accuracy.}%
\label{tab:mistral234}
\begin{tabular}{lcc|cc|cccccc}
 \toprule
  \bf{Size} & \bf{Method} & \bf{Avg bits} & \bf{Wiki2$\downarrow$} & \bf{C4$\downarrow$} & \bf{WinoGrande$\uparrow$} & \bf{PiQA$\uparrow$} & \bf{HellaSwag$\uparrow$} & \bf{ArcE$\uparrow$} & \bf{ArcC$\uparrow$} & \bf{Average accuracy$\uparrow$}\\
  \midrule
  
\multirow{3}{*}{2-bit} & -- & 16.00 & 4.77 & 5.71 & 73.64 & 80.47 & 61.15 & 78.87 & 49.23 & 68.67\\
  & AQLM & 2.01 & 6.32 & 6.93 & 68.75 & 76.01 & 52.13 & \textbf{73.65} & \textbf{40.44}  & 62.17 \\
  & QuIP\# &  2.01 & \textbf{6.02} & \textbf{6.84} & \textbf{69.30} & \textbf{76.71} & \textbf{52.95} & 72.14 & 39.76 & \textbf{62.20} \\
  \cmidrule{2-11}
   & AQLM$^{\bigstar}$ & 2.01 & 5.76 & 6.60 & 68.67 & 77.64 & 56.44 & 73.32 & 42.66 & 63.75 \\

 \midrule
  \multirow{2}{*}{3-bit} & -- & 16.00 & 4.77 & 5.71 & 73.64 & 80.47 & 61.15 & 78.87 & 49.23 & 68.67\\
  & AQLM & 3.04 & \textbf{5.02} & \textbf{5.93} & \textbf{73.24} & \textbf{79.22} & \textbf{59.31} & \textbf{78.28} & \textbf{46.76} & \textbf{67.36} \\
  \cmidrule{2-11}
   & AQLM$^{\bigstar}$ & 3.04 & 5.12 & 6.09 & 72.85	 & 79.05 & 59.92 & 77.57 & 48.12 & 67.50 \\
\midrule
  \multirow{4}{*}{4-bit} & -- & 16.00 & 4.77 & 5.71 & 73.64 & 80.47 & 61.15 & 78.87 & 49.23 & 68.67\\
  & AQLM & 4.02 & 4.89 & 5.81 & 73.80 & 79.71 & 60.27 & 77.86 & 48.21 & 67.97 \\
  & QuIP\# & 4.01 & \textbf{4.85} & \textbf{5.79}  & \textbf{73.95} & \textbf{80.41} & \textbf{60.62} & \textbf{78.96} & \textbf{49.40}  & \textbf{68.67} \\

 \bottomrule
\end{tabular}

\end{table*}

\subsection{Mistral}
Finally, we evaluate AQLM and QuIP\# quantization on Mistral 7b \cite{jiang2023mistral} model for 3 and 4 bits
in Table~\ref{tab:mistral234}. In 2 bits, QuIP\# slightly outperform AQLM on most benchmarks. And for 4 bits setup results are very close across the board.

\section{Pareto optimality}\label{app:pareto_optimality}
\begin{figure*}[!t]
   \begin{minipage}[b]{0.48\textwidth}
     \centering
     \includegraphics[width=\textwidth]{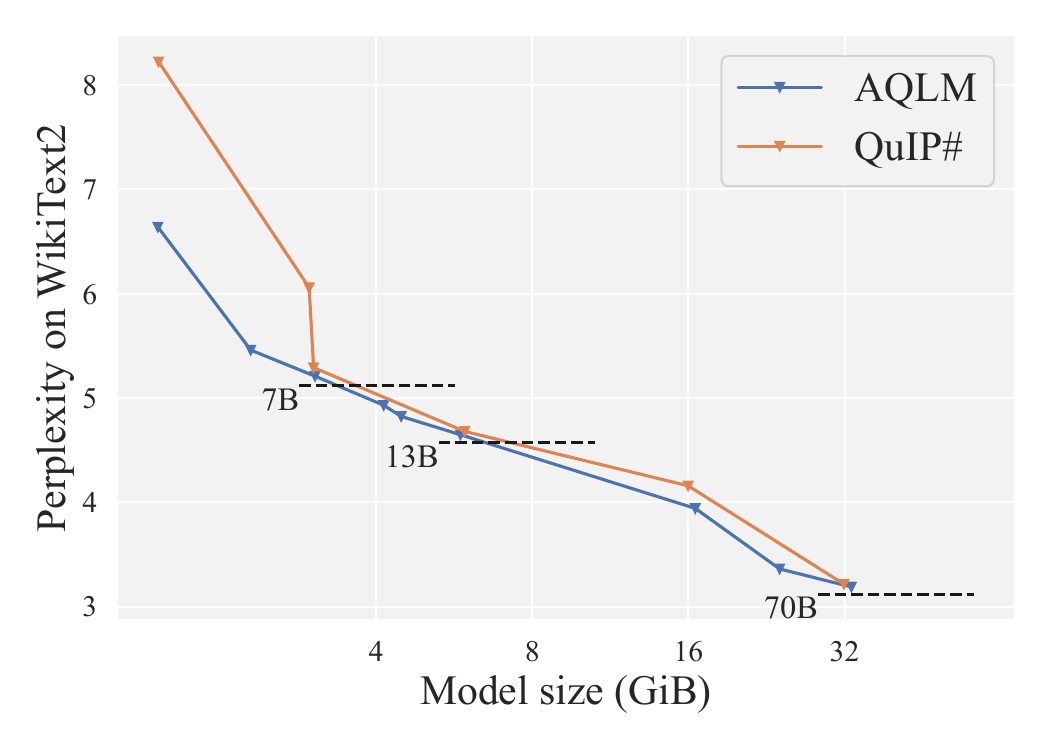}
     \captionof{figure}{%
       Comparison of AQLM relative to QuIP\# on \textsc{Llama 2} 7B, 13B, and 70B models.
     }
     \label{fig:method_comp}
   \end{minipage}\hfill
   \begin{minipage}[b]{0.48\textwidth}
     \centering
     \includegraphics[width=\textwidth]{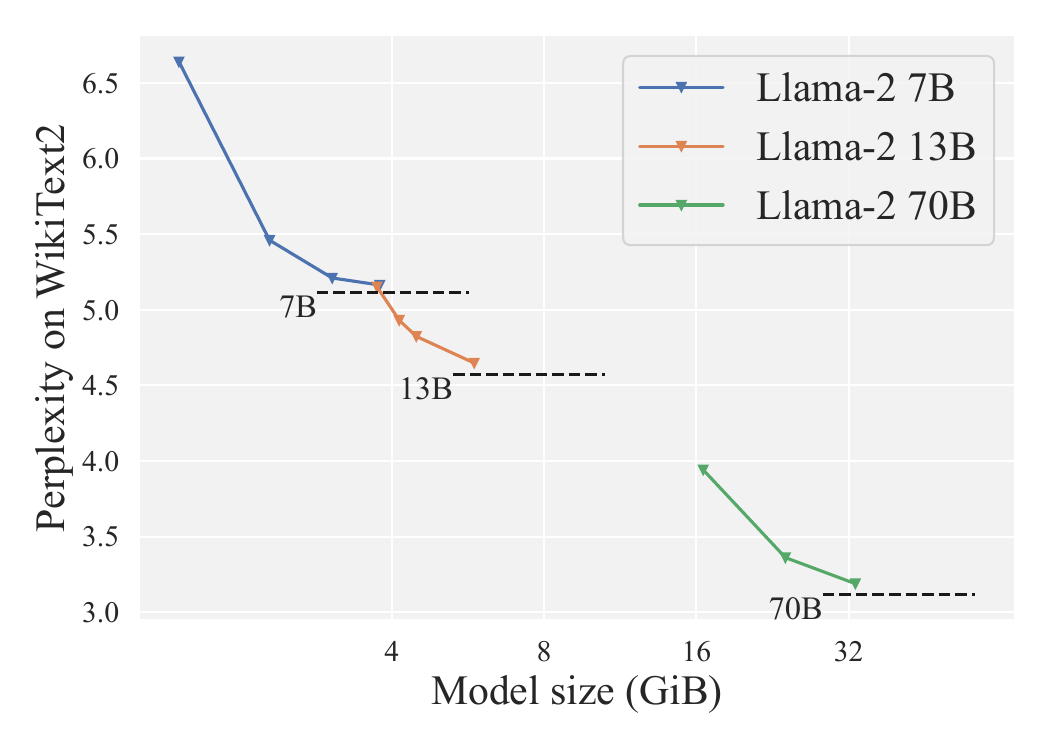}
     \captionof{figure}{%
       Model optimality for AQLM on \textsc{Llama 2} 7, 13, and 70B models.
     }
     \label{fig:model_comp}
   \end{minipage}
 \end{figure*}

We visualize WikiText-2 perplexity of Llama-2 7B, 13B, 70B models quantized with AQLM and QuIP\# as plotted against quantized weight size in bytes and report it in Figure~\ref{fig:method_comp}.
Our method outperforms QuIP\# in terms of perplexity in WikiText-2 across all model sizes.

Additionally, in Figure~\ref{fig:model_comp}, we show perplexity on WikiText-2 for AQLM method against size of quantized parameters. We can notice that starting around 3.7 GiB of quantized weights, which correspond to 2.5 bits compression on \textsc{Llama 2} 13B model, it is more advantageous to compress 13B model rather 7B model at the same model size in bytes.

\section{Estimating model size}\label{app:model_size}

In this section, we describe how to estimate the size of the quantized model for a given codebook configuration.
The total cost of storing quantized weight comprises the codebooks, codes and per-unit scales. 
Specifically for a weight with input dimension $d_{in}$, output dimension $d_{out}$, group size $g$,
$M$ codebooks corresponding to $B$-bit codes, the total amount of memory required is (assuming that codebooks and scales are stored in half precision):
\begin{itemize}
    \item codebooks: $g \cdot M \cdot 2^{B} \cdot 16$
    \item codes: $d_{out} \cdot (d_{in} / g) \cdot B$ 
    \item scales: $d_{out} \cdot 16$
\end{itemize}
Therefore, the \emph{average bits per parameter} can be computed as follows:
\begin{equation}
\bar{b} = \frac{\mathrm{size \ in \ bits}}{\mathrm{number \ of \ parameters}} = 
\frac{16 \ g \ M \ 2^{B} + d_{out} \ (d_{in} / g) \ B \ M + 16 \ d_{out}}{d_{out} d_{in}}
\end{equation}
For example, for \texttt{mlp.gate\_proj} layer of \textsc{Llama 2} 70B model 
with $d_{in} = 8192$, $d_{out} = 28672$, quantization with group size $8$,
two $8$-bit codebooks the formula above yields $2.002$ bits per parameter. Typically, storage cost is dominated 
by the codes, whereas codebooks and scales induce small memory overhead. 

\section{End-to-End Inference Speed}\label{app:generation}

\noindent \begin{minipage}{0.50\textwidth}
    \vspace{0.2cm}
    
    \vspace{-3px}
    \captionof{table}{Text generation speed benchmark.} 
    \label{tab:generation_speed}
    \vspace{-10px}
    \begin{center}
        \begin{tabular}{lcccc}
            \toprule
            \bf{Llama 2} & \bf{7B} & \bf{13B} & \bf{70B} \\
            \midrule
            \multicolumn{4}{c}{Inference on Nvidia RTX 3090 GPU, tok/s} \\
            \midrule
            Original (float16)                    &  54.2    &  29.5      & 5.8      \\
            AQLM ($1\times$16-bit)      &  65.3    &  34.1      & 6.7      \\
            AQLM ($2\times$8-bit)                 &  114.1    &  68.1      & 14.3     \\
            \midrule
            \multicolumn{4}{c}{Inference on Intel i9 CPU, 8 cores, tok/s} \\
            \midrule
            Original (float32)                    &  3.106    &  1.596      & 0.297      \\
            AQLM ($2\times$8-bit)                 &  6.961    &  4.180      & 0.966      \\
            AQLM ($4\times$8-bit)                 &  6.837    &  4.004      & 0.948      \\
            AQLM ($8\times$8-bit)                 &  5.319    &  3.193      & 0.775      \\

            \bottomrule
        \end{tabular}
    \end{center}

\end{minipage}
\hspace{0.05\textwidth}
\begin{minipage}{0.45\textwidth}
    For quantized \textsc{Llama 2} models, setup described in Section~\ref{sect:experiments_inference}, we measure the time it takes to generate 128 tokens from scratch, performed on compiled computational graphs, \textit{with batch size 1}, and report the average number of generated tokens per second on a single 24GB RTX 3090 GPU, as well as Intel i9 CPU, in  Table~\ref{tab:generation_speed}. Perplexity on  WikiText-2 on these configurations presented at the  Table~\ref{tab:groupsxcodebooks}
\end{minipage}
\vspace{0.1cm}

\section{Codebook and codes distribution}\label{app:code_distribution}

\begin{figure}[!t]
    \centering
    \includegraphics[width=0.25\linewidth]{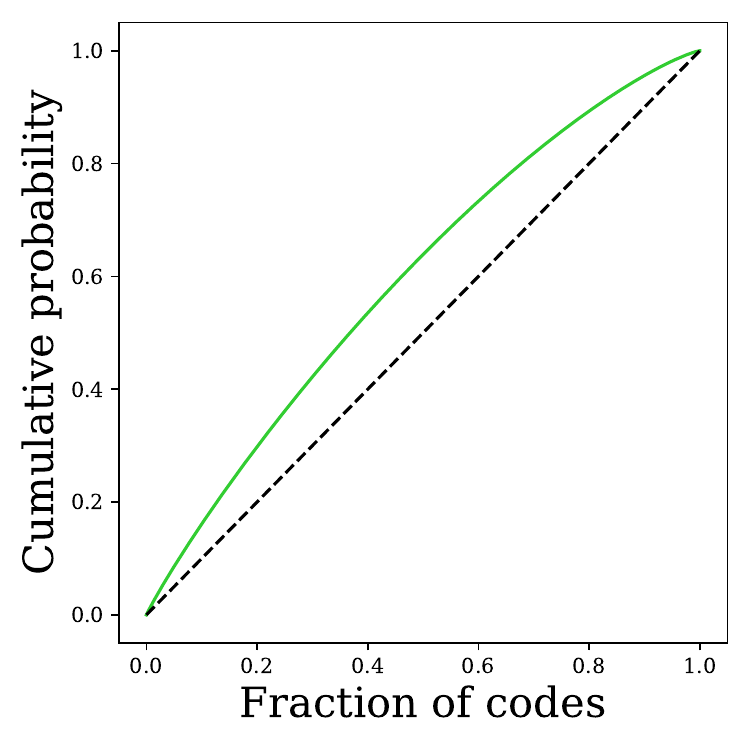}
    \includegraphics[width=0.25\linewidth]{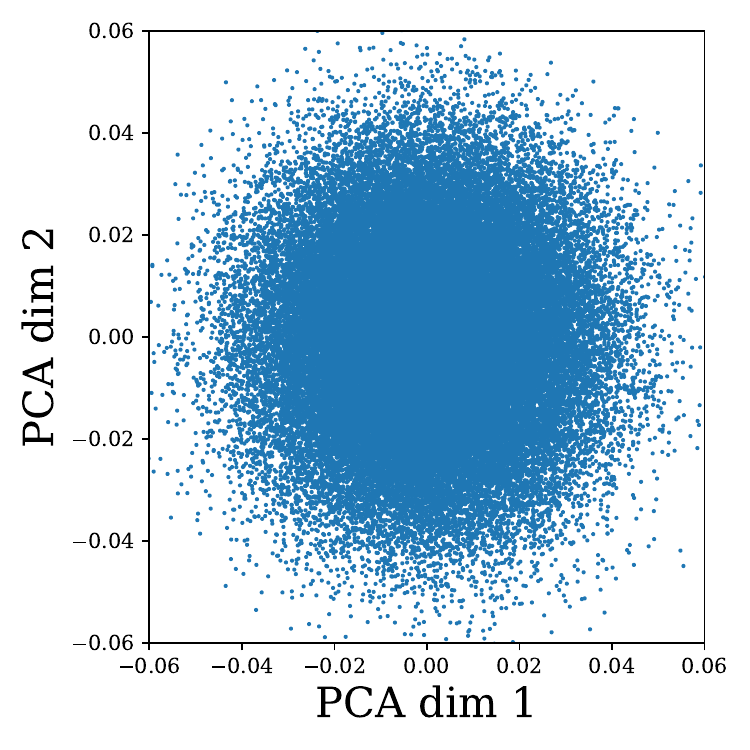}
    \caption{
        Visualization of learned codes and codebooks in \texttt{layers.5.self\_attn.q\_proj}
        linear projection. \\
        (\textbf{Left}) Codes distribution. 
        (\textbf{Right}) Two leading principal components of codebook. 
    }
    \vspace{-5px}
    \label{fig:codes_and_codebooks}
\end{figure}

The proposed AQLM quantization method allows for large freedom in the choice of quantization lattice
and ability to represent different weight distribution. To understand how do the learned codes and codebooks look like,
we visualize the distribution of codes (how frequently given codebook vector is chosen) and the learned codebooks.
Below on Figure \ref{fig:codes_and_codebooks} we provide a cumulative probability plot of leaned codes and two leading principal codebook components for a specific layer. 
One can observe that codes distribution is close to uniform. Its entropy equals 15.91 bits per code, which is close to the maximum possible entropy of 16 bits (for a 16-bit codebook) for the uniform distribution. Codebook vectors are concentrated in some ball. This pattern is pertinent to all linear projections inside transformer blocks. 

\section{Evaluation on MMLU and GSM8k}
\label{app:mmlu_gsm8k_evaluation}

While measurement of perplexity on WikiText-2 and C4 together with zero-shot accuracy on subset of simple 0-shot tasks from LM Eval Harness~\cite{eval-harness} is an established benchmark for evaluation of performance of compressed models, it may be not exhaustive enough for many real-world cases. While the complete and exhaustive evaluation
of LLM abilities is still an open question, we evaluate our AQLM models and QuIP\# on MMLU \cite{mmlu} benchmark that involves problems from 57 different domains, such as humanities, social sciences, physics, e.t.c, and GSM8k \cite{gsm8k} to assess the performance 
of quantized models on more complex and challenging tasks, requiring reasoning to get the correct answer. Below we consider AQLM and QuIP\# 
after end-to-end finetuning, i.e. the best performing quantized models. We observed that relative decrease on performance on these tasks is higher compared to the standard evaluation. Fine-tuned 
AQML and QuIP\# yield very similar performance on these benchmarks. 

\begin{table}[h]
\footnotesize
\centering
\caption{Evaluation of quantized \textsc{Llama 2} models for \textbf{2-2.1 bits per parameter} on MMLU and GSM8k. \\ $^{\bigstar}$ corresponds to end-to-end finetuning}\label{tab:mmlu_gsm8k_evaluation}
\begin{tabular}{lcc|cc}
\toprule
\bf{Size} & \bf{Method} & \bf{Avg bits} & \bf{MMLU (5-shot)} & \bf{GSM8k (8-shot)} \\
\midrule
\multirow{3}{*}{7B} & -- & 16 & 45.9 & 14.6 \\
& QuIP\#$^{\bigstar}$ & 2.02 & 36.8 & \bf{6.2} \\
& AQLM$^{\bigstar}$ & 2.02 & \bf{38.5} & 5.3 \\
\cmidrule{2-5}
\multirow{3}{*}{13B} & -- & 16 & 55.2 & 24.3 \\
& QuIP\#$^{\bigstar}$ & 2.01 & \bf{50.0} & \bf{14.0} \\
& AQLM$^{\bigstar}$ & 1.97 & 48.8 & 13.8 \\
\cmidrule{2-5}
\multirow{3}{*}{70B} & -- & 16 & 68.8 & 56.3 \\
& QuIP\#$^{\bigstar}$ & 2.01 & \bf{65.3} & 46.4 \\
& AQLM$^{\bigstar}$ & 2.07 & \bf{65.3} & \bf{47.9} \\
\bottomrule
\end{tabular}
\end{table}

\section{Block-wise tuning for scalar quantization}
\label{app:block_tuning_gptq}

The block-wise procedure introduced in our work is quite general 
and can be applied to scalar quantization as well. 
Specifically, operations with quantized weights are differentiable with respect to quantization scales kept in original precision. 
Therefore, scales can be tuned in the same way as AQLM codebooks. We observed that tuning significantly improves the quality of GPTQ at low bit widths. However, the resulting quality is still far below 
AQLM at similar bit-widths.

\begin{table}[h]
\footnotesize
\centering
\caption{Evaluation of AQLM and GPTQ quantization after block tuning for \textsc{Llama 2} models with \textbf{2-2.1 bits per parameter}.}\label{tab:gptq_vs_aqlm}
\begin{tabular}{lcc|cc}
\toprule
\bf{Size} & \bf{Method} & \bf{Avg bits} & \bf{Wiki2$\downarrow$} & \bf{C4$\downarrow$} \\
\midrule
\multirow{3}{*}{7B} & -- & 16 & 5.12 & 6.63 \\
& GPTQ & 2.14 & 16.77 & 17.53 \\
& AQLM & 2.02 & \bf{6.64} & \bf{8.56} \\
\bottomrule
\end{tabular}
\end{table}

\end{document}